# Data-driven Traffic Simulation: A Comprehensive Review

*Di Chen, Meixin Zhu\*, Hao Yang, Xuesong Wang, Yinhai Wang*

*Abstract*—Autonomous vehicles (AVs) have the potential to significantly revolutionize society by providing a secure and efficient mode of transportation. Recent years have witnessed notable advancements in autonomous driving perception and prediction, but the challenge of validating the performance of AVs remains largely unresolved. Data-driven microscopic traffic simulation has become an important tool for autonomous driving testing due to 1) availability of high-fidelity traffic data; 2) its advantages of enabling large-scale testing and scenario reproducibility; and 3) its potential in reactive and realistic traffic simulation. However, a comprehensive review of this topic is currently lacking. This paper aims to fill this gap by summarizing relevant studies. The primary objective of this paper is to review current research efforts and provide a futuristic perspective that will benefit future developments in the field. It introduces the general issues of data-driven traffic simulation and outlines key concepts and terms. After overviewing traffic simulation, various datasets and evaluation metrics commonly used are reviewed. The paper then offers a comprehensive evaluation of imitation learning, reinforcement learning, deep generative and deep learning methods, summarizing each and analyzing their advantages and disadvantages in detail. Moreover, it evaluates the state-of-the-art, existing challenges, and future research directions.

## I. INTRODUCTION

D EVELOPING autonomous vehicles (AVs) has become a prevailing trend all around the world due to their considerable advantages and lasting influence on

This study is supported by the National Natural Science Foundation of China under Grant 52302379, Guangzhou Basic and Applied Basic Research Project 2023A03J0106, and Guangzhou Municipal Science and Technology Project 2023A03J0011. Corresponding author is Meixin Zhu (e-mail: meixin@ust.hk)

Di Chen is with the Systems Hub, The Hong Kong University of Science and Technology (Guangzhou) (e-mail: dichen@hkust-gz.edu.cn).
Meixin Zhu is with the Systems Hub at the Hong Kong University of Science and Technology (Guangzhou), the Civil and Environmental Engineering Department at the Hong Kong University of Science and Technology and Guangdong Provincial Key Lab of Integrated Communication, Sensing and Computation for Ubiquitous Internet of Things (e-mail: meixin@ust.hk).
Hao Yang is with the Department of Civil & System Engineering (CaSE) at Johns Hopkins University (e-mail: jhufrankyang@gmail.com).
Xuesong Wang is with School of Transportation Engineering, Tongji University, Shanghai 201804, China and Key Laboratory of Road and Traffic Engineering, Ministry of Education, Tongji University, Shanghai, China. (e-mail: wangxs@tongji.edu.cn).
Yinhai Wang is with the Department of Civil and Environmental Engineering, University of Washington, Seattle, WA 98195 USA (e-mail: yinhai@uw.edu).

future traffic safety and efficiency. With various novel algorithms for AVs surging in recent years, how to conduct efficient and safe tests and validations on these algorithms becomes imperative and challenging.

There are several ways to do AV validation: on-road tests, test track tests, driving simulator experiments, and simulation [1]. They all constitute integral components within the autonomous driving full-chain testing process which is shown in **Fig. 1**.

- On-road tests may initially appear as the most suitable method, but they suffer from inefficiency, high costs, and uncontrollable road conditions, affected by factors such as the traffic environment and the presence of interacting traffic agents. As a result, using immature algorithms or models in on-road tests could compromise the safety of other road users and erode trust in the technology.
- Test track tests are conducted in a controlled environment, which may not accurately reflect real-world driving conditions. Test tracks are designed to simulate different driving scenarios, but they cannot replicate the unpredictable nature of real-world driving, such as unexpected weather conditions, road hazards, and other drivers' behavior.
- Driving simulator experiments may not completely reproduce the real-world driving environment. Driver behavior in such experiments might deviate from that in natural driving situations due to their awareness of being observed [1]. Furthermore, similar to on-road tests and test tracks, driving simulators can be costly and exhibit poor scalability [2].

To address these challenges, intelligent microscopic traffic simulations have been proposed, aiming to make background vehicles in simulated environments react to AV behaviors, in the same way, drivers would do on the road. Microscopic traffic simulation has been studied for decades in the transportation engineering field. However, **bridging the differences in driving behavior between simulation and real world remains a bottleneck** in this field.

There are two main approaches that exist for microscopic traffic simulation: ruled-based and data-driven. Rule-based simulations can generate feasible trajectories without requiring large amounts of data and have strong interpretability, thereby enhancing the credibility of the model and adhering to legal and ethical requirements. However, **they suffer from limited accuracy, poor generalization capability [3], a lack of adaptive updating [4], and the need for significant expert knowledge to establish rules** [5]. Three notable examples of driving simulators are SUMO [6], VISSIM [7], and CARLA [8]. A disadvantage of this solution is that hand-coded actors tend to be unrealistic and rarely present a wide enough variety of behaviors. Alternatively, some driver models, such as the intelligent driver model (IDM) [9], generate more reactive behaviors in response to diverging ego plans. However, their output is limited to deterministic outcomes, ignoring the stochastic nature of vehicle behavior in practical scenarios, and facing challenges in producing diverse scenarios. This is due to the complexity of human behavior, which makes it difficult to describe using manually designed rules.

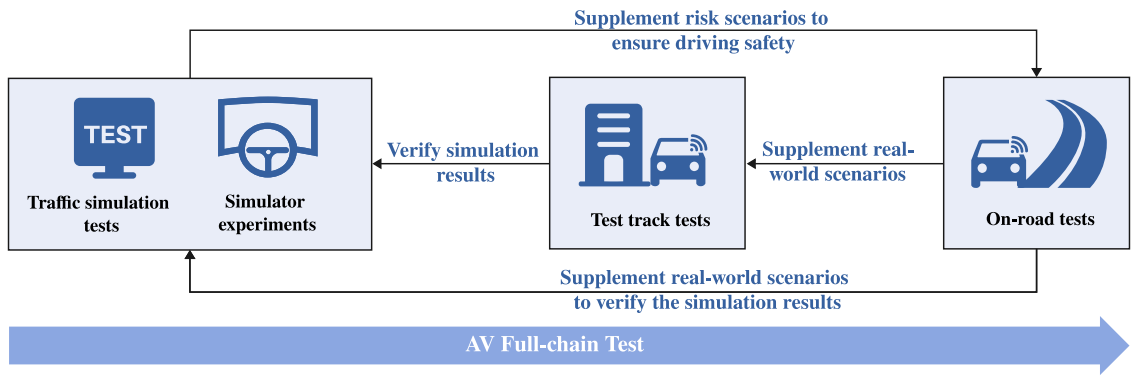

**Fig. 1.** The full-chain test of AV including traffic simulation tests, simulator experiments, test track tests and on-road tests. Traffic simulation tests and simulator experiments provide a cost-effective means of generating long-tail scenarios for the initial testing of AVs. The controlled, real-world environment of track testing can further validate the safety and effectiveness of autonomous vehicles. Subsequently, on-road testing is conducted in uncontrolled and complex traffic environments, which usually represents the final stage of testing.

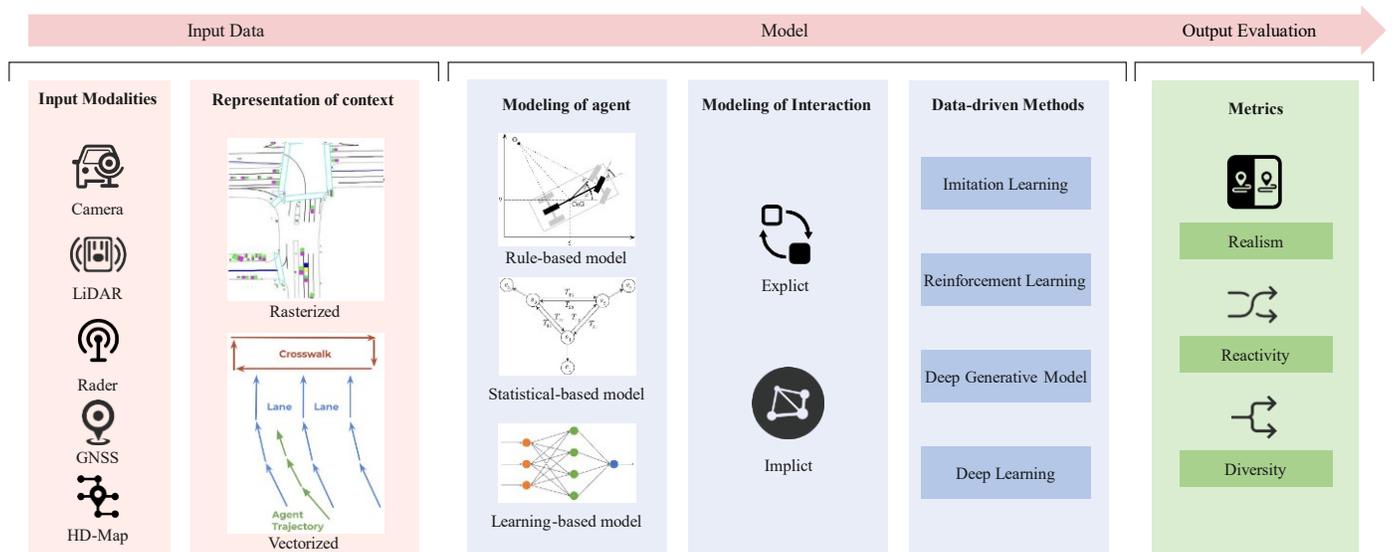

**Fig. 2.** The framework for traffic simulation. Traffic simulation is typically based on the principle of direct perception, in which motion is generated through the direct interpretation of sensory input. To make the data easy to process and understand, the sensory data is often rasterized or vectorized. In the modeling section, agents and their interactions are modeled, followed using data-driven methods for learning. Finally, the output data is analyzed, and the simulation is evaluated from various perspectives using specific metrics.

To address these challenges, data-driven simulations have been proposed to make background vehicles in simulated environments react to AV behaviors in the same way drivers would do on the road. In contrast to rule-based methods, data-driven methods possess distinct advantages. These advantages include the automatic discovery of patterns within substantial datasets, as opposed to manual pattern recognition. Additionally, data-driven methods can generate a wider array of driving behaviors, resulting in greater diversity. Furthermore, they excel in simulating more realistic driving behaviors, particularly in rear-end collision scenarios, which is challenging to capture for expertise. As a result, most research has shifted towards using data-driven methods, which were previously limited by the availability of large amounts of data. Nevertheless, the emergence of many high-quality datasets has weakened this constraint.

This paper presents an extensive review of data-driven traffic simulation methods. The necessity of this review stems from the numerous emerging studies in this field, which warrant a comprehensive and critical analysis. Furthermore, this review addresses a highly relevant subject and fills the existing gap, as no current review sufficiently encompasses the scope and depth of the research question.

The framework of this paper is shown in **Fig. 2**. According to the simulation process, the framework of the article is organized into three main phases: input data, core model, and output evaluation. Traffic simulation is typically based on direct perception, which generates motion directly from sensory input. The simulation system commences by inputting raw sensory data, which is subsequently rasterized and vectorized to enhance the performance of the model. This sensory information is utilized as the input to the backbone model, which is responsible for generating actions, states, or control signals. Finally, the performance of the simulation system needs to be evaluated, usually in terms of realism, reactivity, and diversity.

The contributions of this survey are enlisted as follows:

1. This survey presents an empirical study of data-driven traffic simulation methods. To enhance understanding, we provide a comprehensive introduction to traffic simulation, covering an overview of traffic simulation and data-driven simulation methodology. As far as we are aware, this is the first review of data-driven traffic simulation.

2. An assessment of modeling of agent methods, such as rule-based, statistical-based, and learning-based, analyzing the strengths and weaknesses of each method. And a concise assessment of implicit and explicit interaction modeling.

3. An analytical summary is provided for the metrics and datasets used to evaluate the performance of traffic simulation.

4. A comprehensive evaluation is provided for the prevalent learning models used in traffic simulation, such as imitation learning (IL), reinforcement learning (RL), deep generative, and deep learning (DL) models, along with a discussion of their advantages and disadvantages. Furthermore, future research avenues are identified.

The remainder of this paper is as follows. Section 2 outlines traffic simulation, including the problem formulation, input modalities, dataset, representation of context, modeling of agents, modeling of interaction, and metrics for traffic simulation. Section 3 analyzes the latest developments in data-driven simulation, categorizing them according to IL, RL, generative, and DL methods, and discusses the advantages and limitations of each method. Finally, in Section 4, the state-of-the-art techniques are evaluated, challenges are identified, and future possibilities are explored.

## II. OVERVIEW OF TRAFFIC SIMULATION

In this section, we begin by defining the non-Markov decision process (non-MDP) and MDP problems for traffic simulation, outlining the states and actions of actors. We then provide an overview of the input modalities and datasets utilized in traffic simulation, which form the foundation for a comprehensive understanding of the surrounding environment for vehicles. Additionally, we present two perspectives, Field of View (FOV) and Bird's Eye View (BEV), on the representation of context. To comprehend the behavior of actors, we introduce the modeling of agents and interactions. Finally, we categorize metrics suitable for evaluating simulation results from various perspectives. The framework of this section is shown in **Fig. 3**.

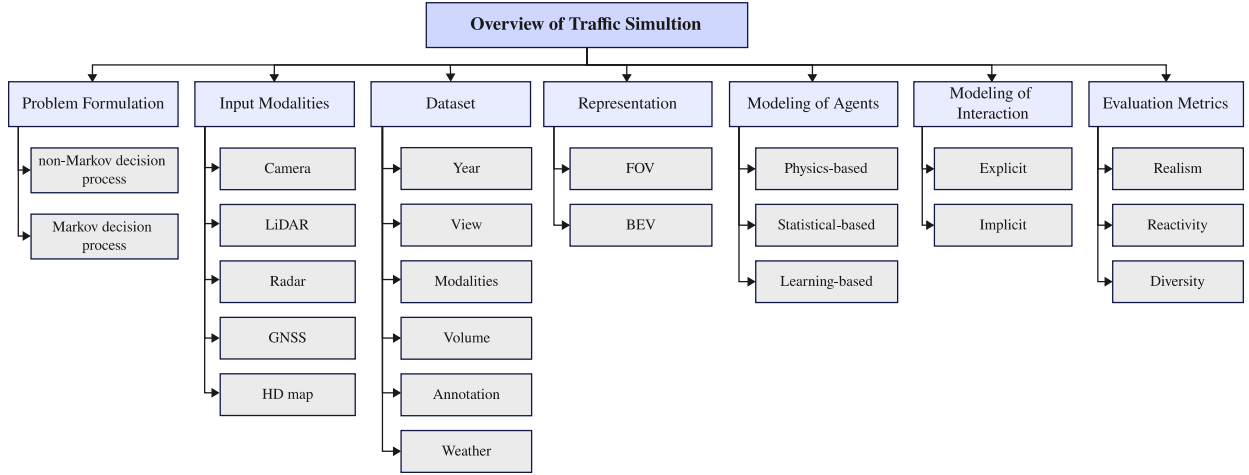

**Fig**. 3. **The framework of the overview of traffic simulation.**

### A. Problem Formulation

The traffic simulation comprises maps and actors. Given a map $M$ and the initial dynamic state $S$ of traffic actors, the objective is to simulate their actions $a$ forward. The map $M$ comprises two components: the static semantic map $M_s$ representing drivable areas such as lanes and intersections, the dynamic environment $M_d$ primarily referring to traffic signals. The state $S$ of $N$ actor states at time $t$ are denoted as $S_t = \{s_t^1, s_t^2, \ldots, s_t^N\}$. The collection of $N$ actor actions at time $t$ is denoted by $a_t = \{a_t^1, a_t^2, \ldots, a_t^N\}$ and is selected from a discrete or continuous range within an action space referred to as *A*. The action space $A$ may encompass high-level commands ("go straight", "turn right", "slow down" etc.), intermediate-level waypoint trajectories for vehicles to track, or low-level control actions such as throttle, brake, and steering. Thus, the key components of traffic simulation can be described by a tuple $\{M, S, A\}$. Based on the above concept, the traffic simulation model can be formulated:

$$a_{T,} \Leftarrow f(M, S_{1:T}, A; \theta) \qquad (1)$$

where $f$ is a behavioral model of actors that takes environmental information, (historical and current) states, and action space as inputs, and outputs for future actions of all controllable actors in traffic scenarios. The argument $\theta$ is a set of parameters in a learning-based model.

Based on the above-mentioned problem definition, traffic simulation can also be modeled as MDPs, as learning models have demonstrated their capability to address sequential decision problems. The goal in solving an MDP problem is to find an optimal policy, which is a mapping from states to actions, that maximizes the expected cumulative reward over time. Each vehicle is an agent that interacts with its environment by making observations and selecting appropriate actions, e.g., steering, and longitudinal acceleration, via a policy $\pi$. This policy helps an agent make decisions that lead to the best possible outcomes in an uncertain environment. In the MDP problem, vehicles are described as agents that can perceive the environment, process information, and take actions to achieve predefined goals or objectives. In addition to the aforementioned elements such as action space $A$ and state space $S$, it also includes elements like the transition probability function $P$, and reward function $R$. So, the MDP problem can be formulated as:

$$M = \{S, A, P_a, R_a, \gamma\} \qquad (2)$$

where $P_a$ denotes the describes the probabilities of transitioning from state $s$ to $s'$ when specific actions $a$ are taken. $\gamma$ is the discount factor. It models how the traffic system evolves over time, considering the effects of actions and uncertainties. $R_a$ assigned numerical values (rewards or costs) received after transitioning from state $s$ to $s'$, due to action $a$. It quantifies the desirability or undesirability of particular traffic conditions and actions, such as minimizing congestion, keeping driving safety, or reducing travel times. When an agent receives incomplete information regarding the state at each time step, the issue can be formally represented as a partially observable Markov decision process (POMDP).

Prior research on traffic modeling has often utilized non-MDP methodologies, which forecast forthcoming events by analyzing past agent trajectories within a brief time frame. This non-MDP framework enables greater adaptability for the network to leverage data across multiple steps, which could potentially improve the precision of predictions. However, these models lacking the MDP structure may present difficulties in solving the sequential decision-making problem under long-time traffic simulation.

*B. Input Modalities*

In traffic simulation, to ensure safe and effective driving decisions, AVs necessitate a detailed and comprehensive understanding of their surrounding environment, encompassing both dynamic objects and static infrastructure. Data for traffic simulation environments are typically derived from multiple input modalities, such as camera, Light Detection and Ranging (LiDAR), radar, Global Navigation Satellite System (GNSS), High-definition (HD) map, etc.

*C. Dataset*

In traffic simulation, the quality and richness of data play a crucial role in model training. The quantity and quality of information provided to the model have a significant impact on the outcomes it produces. We introduce the driving dataset with different views, FOV and BEV. FOV datasets are better suited for training AV algorithms that rely on egocentric information, while BEV datasets are limited to a fixed region, making them more suitable for analyzing the behavior of objects against a stable background. Then, the datasets are summarized based on their data modalities, including camera, LiDAR, GNSS, and HD map. We compared the volume of datasets based on the number of scenes, hours, and frames. Weather conditions can also affect the performance of the model. Some datasets, such as the Waymo open dataset, encompass all weather conditions, ranging from sunny to snowy. The details are provided in TABLE I. It is worth mentioning that the year in the table refers to the most recent year in which the dataset was updated.

*D. Representation of Context*

Two common methods for characterizing autonomous driving simulation environments are through the use of FOV and BEV.

FOV approaches entail observing the world from the vehicles' perspective, typically utilizing visual data such as images or videos as inputs. A key advantage of the FOV approach is its provision of a more realistic and immersive simulation experience, as agents perceive the environment from their own viewpoint [10]. Consequently, this can result in more accurate and realistic agent behavior, as they respond to the environment in a manner that closely resembles real-world driving.

BEV approaches employ algorithms to process data from a top-down, map-like view. In comparison to the FOV counterpart, BEV representations of the world, particularly in traffic scenarios, offer rich semantic information, precise localization, and absolute scales that can be directly utilized by various subsequent applications (e.g., planning, and control) [11].

In BEV, for traffic simulation, we need to learn context representation from the map. There are three main methods exist: rasterized representation, vectorized representation and graph-based representation.

Many researches [12, 13] addressed this problem as a semantic segmentation issue, which rasterizes map elements into pixels and labels each pixel with a class [14]. This formulation enables the direct utilization of fully convolutional networks. Rasterized maps are computationally efficient and straight-forward to process, as they are represented as a pixel grid, which makes them ideal for tasks such as image classification and object detection. Moreover, they can be seamlessly integrated with other image-based data types, such as LiDAR or satellite imagery. However, they have two primary drawbacks. Firstly, they do not offer instance information that is required to distinguish map elements with identical class labels but different meanings, such as left and right boundaries. Secondly, ensuring spatial consistency within the predicted rasterized maps is challenging, as nearby pixels may possess opposing semantics or geometries [14].

The vectorized map represents road structures including traffic signs, lanes, buildings, drivable areas, etc., which are often defined in terms of geometric primitives such as lines, planes, and meshes. Vectorized maps offer the advantage of representing intricate geographic features with a high degree of precision and detail because data isn't dependent on grid size. However, they also have several drawbacks, including their potential for being more computationally intensive and memory-consuming than rasterized maps, particularly when working with large datasets.

Graph-based methods, utilized for representing context, model relationships and dependencies between entities through nodes and edges in a graph. These methods exhibit high levels of expressiveness, robustness, and flexibility in portraying complex relationships and dependencies between various entities. However, the inherent complexity and sparsity of graphs present significant challenges in managing large datasets, often resulting in less accurate representations.

The illustrations of rasterized representation, vectorized representation and graph-based representation are shown in **Fig. 4**.

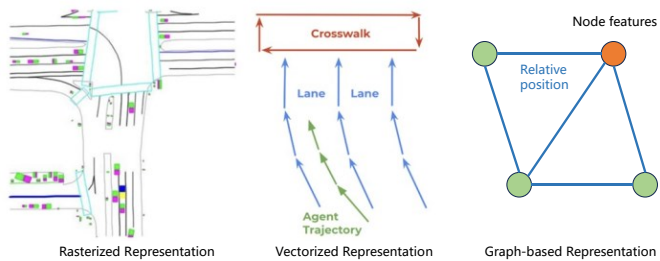

**Fig. 4.** Illustrations of rasterized representation, vectorized representation [15] and graph-based representation [16]. In graph-based representation, the features of the node may include information regarding the type of vehicle, the desired destination, the route points, the traffic lights, and other elements.

### E. Modeling of Agents

The modeling of vehicle agents is to simulate the behavior of vehicles in traffic, which can be achieved through three primary approaches: physics-based, statistical-based, and learning-based. These models consider factors such as vehicle dynamics, road conditions, traffic flow, and driver behavior, with the objective of generating a realistic representation of how vehicles move and interact with other traffic participants in a given scenario.

Physics-based [17] methods refer to all the scientific hypotheses about the movement of cars, taking into account factors such as current position, velocity, acceleration, and road constraints, e.g. IDM, Krauss, Adaptive Cruise Control (ACC), etc. They are transparent and interpretable, which means that the rules used to generate behavior can be easily understood and modified by human experts. What's more, they are computationally efficient, as they do not require large amounts of training data or complex neural network architectures. However, lacking consideration of the anthropomorphic nature of agents and dynamic environmental adaptability, they cannot be applied to situations that are not well-defined.

Statistical-based methods that utilize probabilistic models facilitate the prediction of complex maneuvers, such as lane changes and turns at intersections, by revealing the underlying pattern and accounting for the uncertainty and variation of motion patterns. Many existing works utilized Gaussian processes (GPs) [18-20], Monte Carlo sampling [21, 22], Gaussian mixture models (GMMs) [23], and hidden Markov models (HMM) [24]. However, these models often assume that vehicles are independent entities and don't account for interactions within the context or with other agents. Additionally, they may struggle with complex patterns or long-term dependencies.

Learning-based methods refer to cutting-edge models that mimic human intelligence, mostly leveraging neural networks (NNs) as driving models. The NNs commonly used in research related to this field have been classified into many categories, such as multi-layer perceptron (MLP) [15], convolutional neural network (CNN) [25], recurrent neural network (RNN) [26-28], graph neural network (GNN) [29], transformer, etc. They can be used to generate behavior that is not limited by pre-defined rules or heuristics, which can make them more flexible and adaptable to different scenarios. Additionally, learning-based training methods such as reinforcement learning can be utilized to account for implicit interaction relationships between agents. However, learning-based methods may require large amounts of training data and complex neural network architectures, which can be computationally expensive and time-consuming. Furthermore, learning-based methods may not be as transparent or interpretable as physics-based and statistical-based methods, which can make it difficult to understand how the behavior is generated. Existing learning-based methods are typically insufficient, yielding behaviors of traffic participants that frequently collide or drive off the road, especially over a long horizon. Many researchers [30] have concentrated on exploring methods that combine learning-based approaches with other techniques to enable the generation of interpretable behavior that adheres to dynamics while also demonstrating adaptability to complex environments.

TABLE I
COMPARISON OF DRIVING DATASETS

| Dataset | Year | View | Data Modalities | | | | HD map | Data Volume | | | Annotation | | | Weather |
|---|---|---|---|---|---|---|---|---|---|---|---|---|---|---|
| | | | Camera | LiDAR | Radar | GNSS | | Scenes | Hours | Frames | 3D | 2D | Lane | |
| Waymo (Motion) [31] | 2023 | BEV | ✓ | ✓ | × | ✓ | ✓ | 1K | - | 15M | ✓ | × | ✓ | × |
| Argoverse 2 (LiDAR) [32] | 2022 | BEV | × | × | × | ✓ | ✓ | 25M | 763 | - | × | × | × | × |
| Argoverse 2 (Motion) [32] | 2022 | BEV | ✓ | × | × | ✓ | × | 3 | 6 | - | ✓ | × | ✓ | × |
| InD [33] | 2020 | BEV | ✓ | × | × | ✓ | × | 4 | 16.5 | - | × | ✓ | ✓ | × |
| RoundD [34] | 2020 | BEV | ✓ | × | × | ✓ | × | 4 | 2.5 | - | × | ✓ | ✓ | × |
| INTERACTION [35] | 2019 | BEV | ✓ | ✓ | ✓ | ✓ | ✓ | 366 | 1118 | 4.6M | × | ✓ | ✓ | × |
| Lyft Level 5 [36] | 2019 | BEV | ✓ | ✓ | × | ✓ | ✓ | - | 214 | 2000M | ✓ | ✓ | ✓ | × |
| HighD [37] | 2018 | BEV | ✓ | × | × | ✓ | × | - | 104 | - | ✓ | ✓ | ✓ | ✓ |
| NGSIM[38] | 2007 | BEV | ✓ | ✓ | × | ✓ | ✓ | 2030 | 11.3 | 39M | ✓ | ✓ | ✓ | × |
| Waymo (Perception) [39] | 2023 | FOV | ✓ | ✓ | × | ✓ | ✓ | 2M | - | 6M | × | ✓ | ✓ | ✓ |
| Argoverse 2 (Sensor) [32] | 2022 | FOV | ✓ | ✓ | × | ✓ | × | 1K | 6.4 | 20M | ✓ | ✓ | ✓ | × |
| OpenLane [40] | 2022 | FOV | ✓ | × | × | ✓ | × | - | 2.5 | 5K | ✓ | × | ✓ | ✓ |
| ONCE-3DLanes [41] | 2022 | FOV | ✓ | ✓ | × | ✓ | × | 179 | - | 1.4M | ✓ | ✓ | ✓ | ✓ |
| PandaSet [42] | 2021 | FOV | ✓ | ✓ | × | ✓ | × | - | 144 | 41.7M | ✓ | ✓ | ✓ | × |
| ONCE [43] | 2021 | FOV | ✓ | ✓ | × | × | × | - | - | 21.1M | ✓ | ✓ | ✓ | ✓ |
| A2D2 [44] | 2020 | FOV | ✓ | ✓ | × | ✓ | × | - | - | 1.25M | ✓ | ✓ | ✓ | ✓ |
| Cityscapes 3D [45] | 2020 | FOV | ✓ | ✓ | × | ✓ | × | 11 | - | 8M | ✓ | ✓ | ✓ | × |
| KITTI-360 [46] | 2020 | FOV | ✓ | × | × | ✓ | × | 4 | 10 | - | ✓ | × | ✓ | × |
| HDD [47] | 2019 | FOV | ✓ | ✓ | × | ✓ | ✓ | 113 | 320 | 2.2M | ✓ | × | ✓ | × |
| Argoverse 1 [48] | 2019 | FOV | ✓ | ✓ | ✓ | ✓ | ✓ | 1K | 5.5 | 4M | ✓ | ✓ | × | × |
| nuScenes [49] | 2019 | FOV | × | ✓ | × | ✓ | ✓ | - | 574 | 20M | ✓ | ✓ | ✓ | ✓ |
| Oxford RobotCar [50] | 2019 | FOV | ✓ | ✓ | × | ✓ | × | 160 | 0.8 | 2.8M | ✓ | × | ✓ | ✓ |
| H3D [51] | 2019 | FOV | ✓ | ✓ | × | × | × | - | 55 | 3.9M | ✓ | ✓ | ✓ | × |
| A*3D [52] | 2019 | FOV | ✓ | ✓ | × | ✓ | ✓ | 103 | 2.5 | 14.4M | ✓ | ✓ | × | ✓ |
| ApolloScape [53] | 2018 | FOV | ✓ | × | × | ✓ | × | - | 1100 | 10M | × | ✓ | ✓ | ✓ |
| BDD100K [54] | 2018 | FOV | ✓ | × | × | ✓ | ✓ | 6 | 11.5 | - | × | ✓ | ✓ | ✓ |
| KITTI [55] | 2013 | FOV | ✓ | ✓ | × | ✓ | × | 22 | 1.5 | 1.5M | ✓ | ✓ | ✓ | ✓ |

## F. Modeling of Interaction

Modeling of interaction is an important task in simulation when reasoning about multi-agent behaviors. Many existing approaches rely on using implicit latent variables to model interactions [56-59] or incorporating modeling interaction into agent modeling which reflects interactivity through the reactions of other vehicles. Learning from real-world data, implicit interaction relationships exhibit greater adaptability under complex tasks.

Modeling explicit interaction relations can gain better interpretability. These explicit relations allow to production and manipulation of different types of interactive scenarios. In prior studies [60-62], agent relations have been defined based on pass and yield relationships and predicted using a machine learning model as a classification problem. These predicted interactions help guide multiple agents to generate consistent trajectories [63].

In addition to the above methods, one of the most popular approaches is the game-theory-based framework in decision-making for AVs that consider interaction. However, game theory models are typically designed for specific traffic scenarios such as intersections without traffic lights and lane-changing, lacking universality. Therefore, learning-based implicit or explicit interaction modeling methods are still mostly used in traffic microsimulation.

## G. Evaluation Metrics for Traffic Simulation

Simulation evaluation is commonly classified into two types: open-loop and closed-loop evaluation. In open-loop evaluation, the simulation system is designed to mimic a human driver. During open-loop evaluation, the agent's predictions are not utilized to drive it forward, which implies that the network never receives its own predictions as input [64]. In this case, no agent interactions are considered. In closed-loop evaluation, the simulator generates a planned trajectory using the available information at each timestep, similar to open-loop evaluation [65]. However, in closed-loop evaluation, the proposed trajectory is utilized as a reference for a controller, and as a result, the planning system is gradually corrected at each timestep with the updated state of the vehicle. Furthermore, closed-loop evaluations are crucial for accurately assessing the real performance of driving models, as open-loop evaluations can be misleading [64]. Moreover, evaluation indicators for open-loop and closed-loop evaluation often overlap.

Evaluation metrics are crucial in traffic simulation as they provide a quantitative measure of the realism, reactivity, and diversity of the simulated traffic system. These metrics are specifically designed to assess various aspects of the simulation and can be utilized to compare the simulated results with real-world data or other simulation models. The evaluation metrics applied in traffic simulations are shown in TABLE II.

***Simulation realism***: Simulation realism metrics refer to the measure of the difference between the simulation output and input data. An ideal simulation system should be able to accurately reproduce the behavior of other agents recorded in the input data. They can measure how well the simulation can reconstruct real-world traffic scenarios. In some literature, they are also defined as controllability or reconstruction ability indicators. Agents that possess controllability or reconstruction ability should be able to generate trajectories or behaviors that approximate the distribution of real-world data.

***Simulation reactivity:*** Simulation reactivity metrics are utilized to assess the capacity of target vehicles to respond safely and effectively to the dynamic and intricate transportation environment. It is important to note that complex traffic conditions go beyond hazardous situations, as reactivity indicators encompass safety evaluations throughout the driving process.

***Simulation diversity***: Simulation diversity metrics evaluate the discriminability and the coverage of agent policies. Higher diversity is desired, but it is only meaningful when the diverse samples have a similar level of realism.

## III. DATA-DRIVEN SIMULATION METHODOLOGY

Data-driven methods employ large datasets and machine learning techniques to explore, discover, and analyze implicit relationships between vehicle behavior and the surrounding environment within the data, and to utilize this information for tasks such as trajectory prediction and intent recognition. It aims to generate varied and credible behaviors by leveraging real-world driving logs as demonstrations. From a modeling perspective, data-driven methods can be classified into four major types: IL models, RL models, deep generative models, and DL models. In this section, reviews of the three models are provided, respectively, as shown in **Fig. 5**. The recent methods in data-driven traffic simulation can be seen in TABLE III.

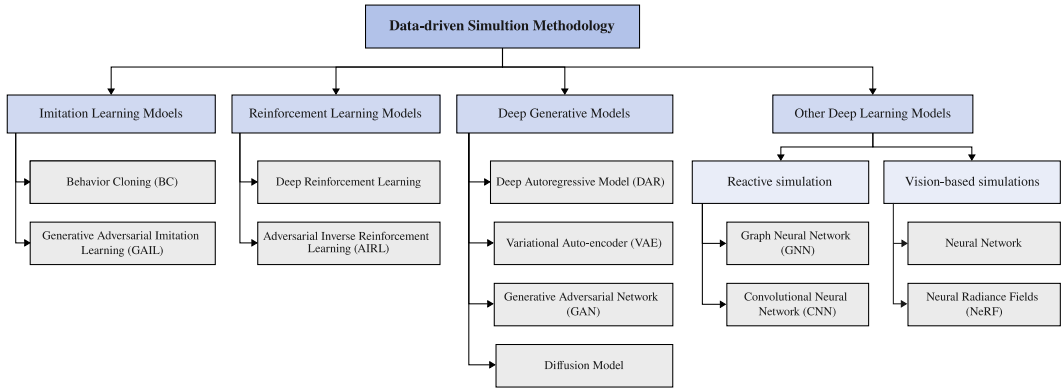

**Fig. 5. The framework of data-driven simulation Methodology**

TABLE II
EVALUATION METRICS APPLIED IN THE LITERATURE

| Perspectives | Metrics | Formula | Description |
|---|---|---|---|
| Realism | Final Displacement Error, FDE | $FDE = \left\| \hat{Y}_{end} Y_{end} \right\|$ | The discrepancy between the predicted final location $\hat{Y}_{end}$ and the actual final location $Y_{end}$ at the conclusion of the prediction horizon. |
| | Average Distance Erroe, ADE | $ADE = \frac{1}{T} \sum_{t=1}^{T} \left\| \hat{Y}_t - Y_t \right\|_2$ | the distance between the predicted location $\hat{Y}$ and the actual location $Y_t$ throughout the prediction horizon, which is defined as T |
| | minADE | | The smallest ADE value among K predictions |
| | minFDE | | The smallest FDE value among K predictions |
| | meanADE | | The average ADE value among K predictions |
| | meanFDE | | The average FDE value among K predictions |
| | Along-track Error, ATE | | The lateral distance error between the vehicle's current position and the reference trajectory |
| | Cross-track Error, CTE | | The perpendicular distance error between the vehicle's current position and the reference trajectory. |
| | Miss Rate, MR | | The percentage of missed cases over all cases |
| | Maximum Mean Discrepancy, MMD | $MMD^2(p,q) = \mathbb{E}_{x,x'\sim p}[k(x,x')] + \mathbb{E}_{y,y'\sim q}[k(y,y')] - 2\mathbb{E}_{x\sim p, y\sim p}[k(x,y)]$ | A measure of the distance between two distributions $p$ and $q$ |
| | Jensen-Shannon Divergence, JSD | $JS(P\|Q) = \frac{1}{2}[KL(P\|M) + KL(Q\|M)]$ | A measure of the similarity between two distributions $P$ and $Q$. M is the average of $P$ and $Q$. |
| | Wasserstein distance, WD | $W(P,Q) = \inf_{\gamma \sim \Pi(P,Q)} E_{(x,y)\sim y}[\|x - y\|]$ | A measure of the distance between two distributions $P$ and $Q$. |
| | Negative log-likelihood, NLL | $H(q,p) = E_{x\sim p} - \log(p(x))$ | The trajectory distribution of the model is represented by $p$, whereas $q$ denotes the distribution of the ground truth data. |
| | mean average precision, mAP | $mAP = \frac{1}{N} \sum_{i=1}^{N} AP_i$ | $AP$ is the average precision for $i$ |
| | Root Mean Square Error, RMSE | $RMSE = \frac{1}{T} \sum_{t=1}^{T} \sqrt{\frac{1}{M} \sum_{m=1}^{M} \left(s_t^{(m)} - \hat{s}_t^{(m)}\right)^2}$ | It measures the difference between the recovered data and the actual data. $s_t^{(m)}$ and $\hat{s}_t^{(m)}$ are the true value and recovered value of the position or speed of vehicle $m$ at time $t$ respectively. |
| | Coverage | $Coverage = \frac{\sum_{f=-N}^{N} v_{model}(f)}{\sum_{f=-N}^{N} max(v_{model}(f), v_{human}(f))}$ | $v(f)$ is the amplitude at frequency |
| | Intersection over union, IoU | | The intersection over the union area of model and human data |
| | Fréchet Inception Distance, FID | | The Fréchet inception distance between the synthesized images and the ground-truth images |
| | Fréchet Video Distance, FVD | | The Fréchet video distance between the synthesized and the truth videos. |

TABLE II
EVALUATION METRICS APPLIED IN THE LITERATURE

| Perspectives | Metrics | Formula | Description |
|---|---|---|---|
| Realism | Route Consistency, RC | | It measures the path-to-path distance between an agent's desired route and the route reconstructed from its executed trajectory, averaged over agents and scenarios. |
| Reactivity | Collision Velocity, CV | | The relative speed at the time of collision |
| | Collision rate, CR | $CR = \dfrac{Num\,(\,collision\ vehicles/scenarios/trajectories\,)}{Num\,(\,all\ vehicles/scenarios/trajectories\,)}$ | The average percentage of collision vehicles in each scenario or collision scenarios or trajectories. |
| | Success Rate, SR | | The ratio of episodes where the agent arrives at the destination |
| | Failure Rate | | The average fraction of agents experiencing a critical failure |
| | Off-Road, OR | | The percentage of time that an interactive agent spends off the road |
| | Average Collision Time, ACT | | The average time between the detection of an obstacle and the occurrence of a collision. |
| | Average Collision Frequency Per Second, CPS | | The average number of collisions that occur per second of driving time. |
| | Average Collision Frequency Per Meter, CPM | | The average number of collisions that occur per meter of driving distance. |
| | Average Collision Distance, ACD | | The average distance between the vehicle and the obstacle at the time of collision. |
| | Number of lane changes | | The average of the number of lane changes made by the self-vehicle in each time period. |
| | Traffic Rule Violation Rate, TRV | $TRV = \dfrac{Num(traffic\ rule\ violations)}{Num(all\,traffic\ rule\ observation)}$ | It measures the percentage of time that a vehicle violates traffic rules or regulations during a given period. |
| | Progress | | The total traveled distance of all simulated agents divided by the number of simulated agents. |
| Diversity | Map-aware Average Self-distance , MASD | $MASD = \max\limits_{k,k' \in 1,\ldots,l} \dfrac{1}{NT_{pred}} \sum\limits_{n=1}^{N} \sum\limits_{t=1}^{T_{pred}} \left\| z_t^{n,k} - z_t^{n,k'} \right\|^2$ | The average distance between the two most distinct samples that do not violate traffic rules |
| | Mean Wasserstein distance, meanWD | $meanWD = \dfrac{2}{n(n-1)} \sum\limits_{i=1}^{n-1} \sum\limits_{j=i+1}^{n} \text{Wass}\,(\rho_i, \rho_j)$ | The Wasserstein distance between the density profiles $\rho$ |
| | Inter-policy Diversity, IPD | $D_{IP}(\Pi) = \dfrac{1}{|\Pi|(|\Pi|-1)} \sum\limits_{\pi \in \Pi} \sum\limits_{\pi' \in \Pi} D_{IP}(\pi, \pi')$ <br> where $D_{IP}(\pi, \pi') = \dfrac{1}{|S_\pi \cap S_{\pi'}|} \sum\limits_{s \in S_\pi \cap S_{\pi'}} d\left(\tau_s(\pi), \tau_s(\pi')\right)$ | The average distance of two policy pairs |
| | Overall Diversity, OAD | $D_{OA}^{\mathcal{T}}(\Pi) = \dfrac{1}{|S|} \sum\limits_{s \in S} D_{OA}^{\mathcal{T},s}(\Pi_s)$ <br> where $D_{OA}^{\mathcal{T},s}(\Pi) = \inf\limits_{\gamma \in \Gamma(\tau_s(\Pi),\mathcal{T})} \mathbb{E}_{(\tau,\tau') \sim \gamma}[d(\tau, \tau')]$ | The distance between the obtained trajectory and the expected trajectories |

TABLE III
RECENT METHODS IN DATA-DRIVEN SIMULATION

| Learning Method | Ref. | Method | Year | Dataset | MDP | Representation of context | Backbone Model | Interaction | Evaluation Perspective | Loss |
|---|---|---|---|---|---|---|---|---|---|---|
| IL | Xu, et al. [66] | BITS | 2023 | Lyft Level 5, nuScenes | √ | Raster | ResNet-18 | Implicit | Realism, Diversity | Cross-entropy loss |
| | Guo, et al. [16] | HMMIL | 2023 | pNEUMA | √ | Graph | GNN | Implicit | Realism, Reactivity | NLL loss |
| | Bergamini, et al. [67] | BC | 2021 | Lyft Level 5 | √ | Raster | ResNet-50 | Implicit | Realism, Reactivity | Imitation loss |
| | Zheng, et al. [68] | IL | 2020 | Private dataset | √ | - | NN | Implicit | Realism | Binary classification negative log loss |
| | Yan, et al. [69] | GAIL | 2023 | RoundD | √ | - | Transformer | Implicit | Realism | Adversarial loss, Imitation loss |
| | Zhu, et al. [70] | GAIL | 2023 | NGSIM | √ | - | Diffusion model | Implicit | Realism | Entropy loss |
| | Bhattacharyya, et al. [71] | GAIL | 2022 | NGSIM | √ | - | GRU/MLP | Implicit | Realism | Logistic loss |
| | Behbahani, et al. [72] | Horizon GAIL | 2019 | Video | √ | - | Resnet-101 | Implicit | Realism, Diversity | Cross entropy loss |
| | Bhattacharyya, et al. [73] | GAIL | 2018 | NGSIM | √ | - | RNN | Implicit | Realism | Average cross-entropy loss |
| RL | Cao, et al. [74] | RLHF | 2023 | nuScenes | √ | - | Diffusion model | Implicit | Realism | Average negative log-likelihood |
| | Sackmann, et al. [75] | AIRL | 2022 | Private dataset | √ | Vector | NN | Implicit | Realism, Reactivity | Binary cross entropy loss |
| | Zheng, et al. [25] | IRL | 2022 | Private dataset | √ | - | CNN | Implicit | Realism | IRL loss |
| | Niu, et al. [76] | DRL | 2023 | HighD | √ | - | NN | Implicit | Reactivity | RL loss |
| | Chen, et al. [77] | DRL | 2022 | SPMD | √ | - | CNN | Implicit | Reactivity | Q-learning loss |
| | Zhang, et al. [78] | DRL | 2022 | INTERACTION | √ | Vector | NN | Implicit | Realism, Reactivity, Diversity | Max-margin classification loss+smooth L1 regression loss+RouteLoss; RL loss |
| | Mavrogiannis, et al. [79] | DRL | 2022 | - | √ | - | MLP | Implicit | Reactivity | MSE loss |
| | Kothari, et al. [80] | DRL | 2021 | Lyft Level 5 | √ | Raster | NN/Replay logs | Implicit | Realism, Reactivity | L2 imitation loss |
| | Feng, et al. [81] | DRL | 2021 | SPMD | √ | - | Decision Tree | Implicit | Reactivity | RL loss |
| | Shiroshita, et al. [82] | DRL | 2020 | - | √ | - | CNN | Implicit | Diversity | RL loss |

TABLE III
RECENT METHODS IN DATA-DRIVEN SIMULATION

| Learning Method | Ref. | Method | Year | Dataset | MDP | Representation of context | Backbone Model | Interaction | Evaluation Perspective | Loss |
|---|---|---|---|---|---|---|---|---|---|---|
| DL | Yang, et al. [83] | CNN | 2023 | PandaSet | × | - | CNN | Implicit | Realism | Photometric loss, perceptual loss |
| | Suo, et al. [29] | GNN | 2023 | Argoverse 2, Highway | √ | Vector | GNN | Implicit | Realism, Reactivity | average Huber loss |
| | Chang, et al. [84] | Vectornet | 2022 | INTERACTION | × | Vector | Vectornet | Implicit | Realism, Reactivity | Train loss |
| | Sun, et al. [63] | RNN | 2022 | Waymo Open Motion Dataset | × | Vector, raster | ResNet-50 | explicit | Realism, Reactivity | Cross entropy loss |
| | Chen, et al. [85] | NN | 2021 | UrbanData, Argoverse | × | Graph | PointNet, MLP | Implicit | Realism | The LiDAR loss |
| Generative Model | Feng, et al. [86] | AR | 2023 | Waymo Open Dataset, Argoverse | × | Vector | LSTM+MCG | Implicit | Realism, Reactivity | Binary cross entropy |
| | Tan, et al. [87] | AR | 2021 | ATG4D, Argoverse | × | Raster | ConvLSTM | Implicit | Realism | Average negative log-likelihood |
| | Zhang, et al. [88] | VAE | 2023 | Waymo Open Dataset | × | Vector | Transformer | Implicit | Realism | Smooth L1+KL-divergence+cross-entropy loss |
| | Rempe, et al. [89] | VAE | 2022 | nuScenes | × | Vector | GNN | Implicit | Reactivity | - |
| | Jiao, et al. [90] | Adversarial AE | 2022 | Argoverse | × | Graph | CNN | Implicit | Realism | Smooth L1 loss, negative Log-Likelihood |
| | Suo, et al. [59] | VAE | 2021 | ATG4D | × | Raster | CNN | Implicit | Realism, Diversity | - |
| | Tang, et al. [91] | VAE | 2021 | INTERACTION | × | Vector | MLP | Implicit | Realism | MSE |
| | Krajewski, et al. [92] | VAE | 2019 | High D | × | - | CNN | Implicit | Realism | Mean Absolute Error KL divergence |
| | Krajewski, et al. [93] | VAE, GAN | 2018 | HighD | × | - | NN | Implicit | Realism, Reactivity | MSE, KL divergence |
| | Zhang, et al. [94] | GAN | 2022 | Lyft Level 5 | × | Vector | Graph Attention Network With Transformer | Implicit | Realism, Reactivity | Negative Log-Likelihood, Mean Absolute Error, Mean Square Error |
| | Yin, et al. [95] | GAN | 2021 | Argoverse, INTERACTION | × | - | MLP | Implicit | Reactivity | Negative Log-Likelihood, Mean Absolute Error |
| | Zhong, et al. [96] | Diffusion model | 2023 | nuScenes | × | Raster | Diffusion model | Implicit | Realism, Reactivity | Analytical loss functions based on STL rules |
| | Pronovost, et al. [97] | Diffusion model | 2023 | Private dataset | × | Raster | CNN | Implicit | Realism | Classification+L1+vertex loss, reconstruction loss+L2 |
| | Wang, et al. [98] | Diffusion model | 2023 | nuScenes | × | - | Diffusion model | Implicit | Realism | Mean-squared error and L1 loss |

TABLE IV
A COMPARISON OF BC AND GAIL

| Methods | Advantages | Disadvantages |
|---------|-----------|---------------|
| BC | *Deterministic Policy Learning:* BC typically learns a deterministic mapping from states to actions based on expert demonstrations. This can result in stable and predictable behavior. | *Lack of Adaptability to Novel Situations:* BC tends to perform poorly in situations that deviate significantly from the expert's demonstrations. It may struggle to handle unforeseen scenarios or situations not encountered during training. |
| | *Reduced Exploration Effort:* BC directly imitates the expert's actions, bypassing the need for costly exploration in the environment. This can lead to faster training times and more reliable policies. | *Limited Robustness to Noisy Data:* BC is sensitive to noise in expert demonstrations, which can lead to suboptimal policies. If the expert data contains errors or noise, BC may learn incorrect behaviors. |
| GAIL | *Improved Robustness to Noisy Data:* GAIL is more robust to noise in the expert demonstrations compared to BC. It can better filter out errors or noise in the expert data, leading to more accurate policy learning. | *Potential for Mode Collapse:* GAIL, like other GAN-based methods, can suffer from mode collapse, where the generator fails to capture the full diversity of expert behavior. This can lead to a limited range of generated traffic scenarios. |

## A. Imitation Learning Models

To date, IL has proved its noticeable competence in robot learning tasks, the main idea of which is to use supervised learning with expert data to mimic the behavior of demonstrators. IL-based agents can efficiently learn from demonstration (LfD) to perform tasks as expert-like as possible, demonstrating its outstanding advantages in behavioral reproducibility. In terms of autonomous driving, increasing studies in recent years have utilized IL to enable AVs to behave more like expert human drivers, which is mainly attributed to the large development of real traffic data because they can provide a large number of demonstrations to learn. Therefore, following this idea to make BVs in simulations behave like human drivers becomes possible. The key challenge is that we need to run the system closed loop, where errors accumulate and induce a shift from the training distribution [99]. There are two common ways to perform IL: behavior cloning (BC) and Generative adversarial imitation learning (GAIL). A comparison of them is shown in TABLE IV.

BC aims to reproduce behavior close to expert demonstration, in a supervised fashion, learning the mapping from state to action as a behavior predictor from demonstration data, without requiring agents to interact with the environment. This problem can be expressed as using a training dataset $D$ to estimate strategy $\pi_\theta$ which is trained to predict $a$ under $s$. A classical estimation method is maximum likelihood estimation, so the parameter set $\theta$ can be optimized by:

$$\max_\theta \sum_{(s,a)\in D} \log(\pi_\theta(a|s)) \tag{3}$$

However, this approach can result in causal confusion [72]. Furthermore, behavioral cloning training necessitates a significant amount of demonstration data and is vulnerable to distribution shift [99] caused by the discrepancy between the distribution of the training data and the learned policy's state.

Zheng, et al. [68] are the first to consider the traffic simulation problem as a learning problem, comparing GAIL with BC and car-following model. In ChauffeurNet [64], this shift is mitigated by letting the model learn from synthesized data with perturbed driving trajectories which may cause unfavorable outcomes including collisions and driving off the road, where an additional learning loss is used to punish bad action while encouraging good one. SimNet [67] retains BC but also describes the simulation problem as a Markov Process, in which the authors leverage conditional generative adversarial networks (cGANs) [100] to model the initial distribution of states and ResNet-based CNNs [101] to model behavioral policy of BVs. Both methods alleviate the distribution shift problem by introducing synthetic perturbations to the training trajectories. However, SimNet and ChauffeurNet also reveal another failure case in BC, which is known as causal confusion [72], with an example that a vehicle controlled by Chauffeur-Net will keep waiting at the intersection until the vehicle behind it starts moving. Xu, et al. [66] systematically deconstruct the learning problem into a two-tiered hierarchy, comprising high-level goal inference and low-level goal-conditioned policy, and is trained using a bi-level IL approach. This decomposition holds the promise of bolstering both sample efficiency and behavioral diversity. To generate stable long-term simulations, Guo, et al. [16] proposed a history-masked multi-agent IL method that eliminates the historical trajectory information of all vehicles and introduces perturbations to their current positions during the learning process.

GAIL-based methods aim to uncover the hidden reward function of human driving behavior and acquire the driving policy by maximizing the learned reward. GAIL enabling the agent to interact with the environment during training, can theoretically address the covariate shift of BC in a single-agent context. Behbahani, et al. [72] propose video-to-behavior (ViBe) based on a novel Horizon GAIL as well as an actor-critic approach. This end-to-end framework can directly learn a model from raw video data and output the naturalistic behavior of BVs in a decentralized way. Igl, et al. [102] employ model-based GAIL (MGAIL) along with a hierarchical framework dividing the model into goal generation and goal conditioning, using a parallel beam search to preserve behavioral realism and diversity of BVs. Although many strategies have been used to unearth the behavioral diversity buried in real-world data, due to the existence of long-tail events, the models learned from training data cannot cover all circumstances, i.e., the training data may not record some safety-critical events that rarely happen but do exist in the real world. In this case, the models may output abnormal behavior of agents because of missing references. To overcome this problem, [69] develops a GAIL-based behavioral modeling framework with statistical realism. This framework leverages a Transformer model [103] to learn the interaction among BVs and also considers their behavioral uncertainty, allowing unsafe behaviors of BVs prone to safety-critical events such as collisions and near-misses. Zhu, et al. [70] suggested Realistic Interactive TrAffic flow (RITA) to enhance existing simulators by facilitating high-quality traffic flow for evaluating and optimizing driving strategies. In the traffic generation module, they compared multiple GAIL-based algorithms and found that the multiagent reactive agent (MAGAIL) has the highest safety rate, whereas the multi-modal reactive agent (InfoGAIL) exhibits greater diversity.

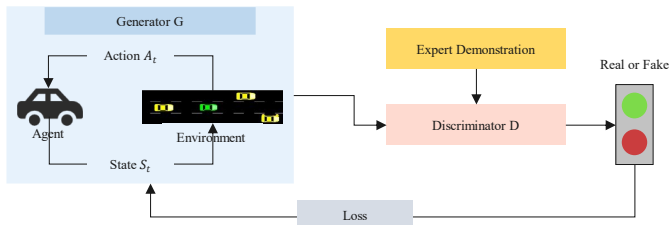

**Fig. 6.** Depiction of the GAIL method. It involves a generator, representing the agent's policy, and a discriminator, which distinguishes between the agent's and expert's actions. During training, the generator and discriminator engage in a minimax game, resulting in an agent that effectively imitates the expert's behavior and performs complex tasks efficiently.

What's more, the performance of GAIL declines when applied to multi-agent IL due to the dynamic environment, resulting in a challenging training process. Bhattacharyya, et al. [73] expand upon GAIL to address the challenge of generalizing models learned from single-agent scenarios to multi-agent driving situations that arise due to observations at training and testing stages being sampled from different distributions. They employed a parameter-sharing technique rooted in curriculum learning to overcome this limitation. Bhattacharyya, et al. [71] proposed variations to GAIL that are useful specifically for the problem of driver modeling: parameter sharing to facilitate multi-agent imitation, reward augmentation to incorporate domain knowledge, and mutual information maximization to reveal individual driving styles. Nonetheless, since the outcome of GAIL is solely a policy and not a reward function [104], it precludes learning in scenarios beyond the training situations.

## B. Reinforcement Learning Models

RL-based methods present a promising solution to tackle distribution shift challenges. The primary goal of RL is to optimize cumulative rewards over time by interacting with the environment, wherein the network derives driving decisions from its actions to acquire rewards or incur penalties. IL, on the other hand, faces difficulties when confronted with novel situations that substantially deviate from the training dataset. Nonetheless, RL demonstrates resilience to this problem as it investigates all pertinent scenarios throughout the training process. Current research focuses on applying Deep Reinforcement Learning (DRL) methods and Adversarial Inverse Reinforcement Learning (AIRL) methods within traffic simulation. A comparison of the two methods when applied in traffic simulation is shown in TABLE V.

Recently, several approaches proposed to apply RL in simulation learning problems such as generating scenarios that match diversity and authenticity or learning a variety of realistic driving skills. Shiroshita, et al. [82] balanced diversity and driving skills by utilizing the representational and exploratory capabilities of DRL. Kothari, et al. [80] proposed an open-source environment, DiverGym, which supports reactive agent behavior and log replay to control agent behavior. Zhang, et al. [78] proposed TrajGen, a method that decomposes trajectory generation into two stages: trajectory prediction, which is generated using LanGCN, and trajectory modification, which is performed to avoid collisions using Twin Delayed Deep Deterministic Policy Gradient algorithm (TD3). They developed a simulator, I-Sim, to ensure the trajectories generated by TrajGen adhere to vehicle kinematic constraints. Mavrogiannis, et al. [79] proposed a traffic simulator that integrates behavior-rich vehicle trajectories associated with varying levels of aggressiveness, utilizing the CMetric algorithm [105]. Subsequently, they employed DRL to train a behaviorally-guided policy, which maps a state to a high-level vehicle control command. Niu, et al. [76] customized a Reversely Regularized Hybrid Offline-and-Online ((Re)$^2$H2O) RL approach to further penalize Q-values on real-world data and reward Q-values on simulated data, which ensures that the generated scenarios are both diverse and adversarial. Feng, et al. [81] constructed a driving environment that offers spatiotemporally continuous testing scenarios for AVs. They addressed the bias and inefficiency issues associated with the Naturalistic Driving Environment (NDE) by employing the importance sampling theory to a small subset of variables while applying the Crude Monte Carlo theory to the remaining variables. In [77], the authors employed DDPG to train adversarial agents and subsequently implemented a nonparametric Bayesian method for clustering the adversarial policies, and ultimately facilitated the generation of vehicle trajectories.

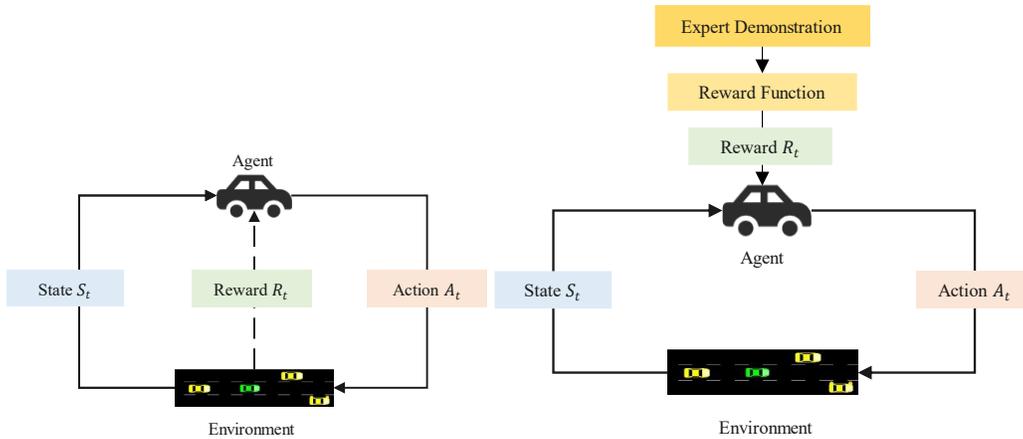

**Fig. 7.** Depiction of the RL (left) and IRL(right) methods. RL methods and IRL methods empower agents to interact with the environment to make decisions and optimize their actions. By receiving feedback in the form of rewards or penalties, the agent seeks to maximize its cumulative reward over time. The key distinction between these approaches lies in the construction of the reward function: in RL, the function is artificially designed, while in IRL, it is derived from the observed behavior of experts.

TABLE V
A COMPARISON OF DRL AND AIRL

| Methods | Advantages | Disadvantages |
|---|---|---|
| DRL | *Adaptability*: DRL can adapt to dynamic environments in traffic simulations, by continually updating its policy based on interactions with the environment. | *Weak robustness:* DRL-based methods can be sensitive to changes in the environment, such as changes in the reward function or the introduction of new obstacles. |
| | *Real-time decisions*: DRL can make real-time decisions, which is essential for simulating traffic scenarios that require quick responses to changing conditions. | *Sparse rewards*: In traffic simulations with sparse rewards, DRL might struggle to optimize its policy due to the lack of sufficient feedback. |
| | *Flexibility*: DRL supports the exploration of various strategies and can learn optimal policies in different traffic situations. | *Sensitivity to hyperparameters*: Hyperparameter tuning can be challenging in DRL, especially in complex traffic simulations with multiple interacting agents. |
| AIRL | *Expert demonstrations*: AIRL can capitalize on expert demonstrations, including real-world traffic data, to expedite the learning process and generate more accurate traffic simulations. | *Dependence on demonstrations*: AIRL relies on the availability of expert demonstrations, which may be difficult to obtain or may not cover all possible traffic scenarios. |
| | *Transferability*: AIRL learns a reward function, which can potentially make it easier to transfer the learned knowledge to other traffic scenarios or environments. | *Mode collapse*: AIRL can suffer from mode collapse, where the learned policy may only cover a subset of expert behavior, resulting in less diverse traffic simulations. |
| | *Robustness to noise*: AIRL is more robust to noise in expert demonstrations, making it suitable for traffic simulations where the data might be noisy or incomplete. | *Computational complexity*: AIRL's adversarial training process for learning both the policy and the reward function can be computationally demanding, especially for large-scale traffic simulations. |

For these above works, the reward design is necessary but complicated because it is difficult to mathematically interpret the vehicle's true objective. So, some researchers focused on AIRL which can reconstruct reward function from expert experience. Zheng, et al. [25] proposed a parameter-sharing AIRL model for dynamics-robust simulation learning which can imitate a vehicle's trajectories in the real world while concurrently recovering the reward function that discloses the vehicle's true objective, which remains invariant to differing dynamics. Sackmann, et al. [106] employed AIRL to reconstruct rewards using a real-world dataset and subsequently maximize these rewards, effectively mitigating distribution shift issues. Upon comparing AIRL with GAIL and BC, they observed that AIRL demonstrates significantly greater robustness.

Incorporating realistic traffic models that align with human knowledge is a crucial component of effective simulation. However, the limited expressiveness of the above methods in capturing human preferences for realism in traffic simulations remains a challenge. Cao, et al. [74] introduce TrafficRLHF, a framework utilizing human feedback to enhance traffic models for realistic scenario generation.

## C. Deep Generative Models

Data-based simulators can implicitly learn multi-vehicle interactions from natural driving environments, and simulate how the vehicles interact with each other. However, these simulations rely heavily on the collected dataset, making it difficult to edit and expand the scenarios, resulting in the problem of scenario fragmentation.

Deep generative models (DGMs) identify probabilistic models that represent the distribution of training data and can generate new samples from the corresponding distribution. In traffic simulation problems, BVs' behavior is assumed to conform to certain probability distributions. Therefore, DGMs are commonly utilized to learn a probabilistic distribution over traffic scenes and are frequently employed for generating long-tail or realistic scenarios. Some common types of DGMs include deep Autoregressive Model (DAR) [86, 87], Variational Auto-encoder (VAE) [88, 89, 91], Generative Adversarial Network (GAN) [67] [94], Diffusion Model [96, 97]. Once the DGMs are successfully trained, they will turn into behavioral models and can be deployed to sample new behaviors of BVs. A comparison of these methods when applied in traffic simulation is shown in TABLE VI.

TABLE VI
A COMPARISON OF GENERATIVE MODELS

| Methods | Advantages | Disadvantages |
|---------|------------|---------------|
| DAR | *Temporal Context Sensitivity:* DAR models are capable of incorporating a rich temporal context, which is crucial for modeling time-dependent traffic patterns and predicting future states. | *Computational Intensity:* Implementing and training DAR models can be computationally demanding, particularly for large-scale traffic networks, potentially requiring significant computational resources. |
| | *Non-Linear Relationships:* DAR models can capture complex non-linear dependencies within traffic data, allowing for a more accurate representation of dynamic traffic behaviors. | *Model Interpretability:* The high complexity of neural networks may make it more challenging to interpret the learned relationships between variables compared to simpler models. |
| VAE | *Latent Space Representation:* VAEs provide a compact latent space representation of traffic data, allowing for meaningful feature extraction, anomaly detection, and traffic segmentation. | *Latent Space Ambiguity:* Interpreting the latent space can be challenging, as individual dimensions may not have clear semantic meanings. Understanding the learned representations may require additional effort. |
| | *Generative Capabilities:* VAEs can generate synthetic traffic samples from the learned latent space distributions, facilitating the creation of new scenarios for micro-traffic simulation. | *Mode Collapse:* VAEs may struggle to capture the full diversity of complex traffic patterns, potentially leading to the generation of samples that lack variability. |
| GAN | *Realistic Sample Generation:* GANs excel at producing highly realistic and diverse traffic samples, which is essential for simulating complex and dynamic traffic environments. | *Mode Collapse:* Like VAEs, GANs can suffer from mode collapse, where the model generates a limited range of traffic scenarios, potentially leading to unrealistic simulations. |
| Diffusion model | *Capturing Complex Dependencies:* Diffusion models excel at modeling intricate dependencies between variables in traffic data, allowing for a more accurate representation of traffic dynamics. | *Interpretability Challenges:* The complex nature of diffusion models may make it more challenging to interpret the learned relationships between variables, requiring additional analytical effort. |

1) DAR: The autoregressive model generates an explicit density model that is facile to manipulate for maximizing the likelihood of the training data. Since transportation environments are highly dynamic, nonlinear and complex, deep autoregressive generative models are often used. Tan, et al. [87] presented a neural autoregressive model named SceneGen, which inserts actors (vehicles) of diverse classes into the scene and generates their sizes, orientations, and velocities. These methods enable the modeling of high-dimensional features of scenarios, thereby enhancing their fidelity and scalability. In Feng, et al. [86] study, TrafficGen employs an autoregressive neural generative model with an encoder-decoder architecture that utilizes a novel vector-based context representation, multi-context gating (MCG) [28], learning to synthesize realistic traffic scenarios from the fragmented data collected from the real-world.

2) VAE: VAE can learn the latent representation of data and generate new data. It is a neural network-based model that consists of an encoder and a decoder. Although VAEs can model diverse behaviors without dropping modes, they are not well-suited for learning robust policies. Krajewski, et al. [93] proposed two neural network architectures, Trajectory Generative Adversarial Network (TraGaN) and Trajectory Variational Autoencoder (TraVAE), capable of generating lane change maneuver trajectories. They demonstrated that TraGaN's latent space is more disentangled and intuitive. Krajewski, et al. [92] introduced a novel neural network architecture, Bézier Variational Autoencoder (BézierVAE), based on TraVAE, that employs a Bézier-curve layer and additional loss terms to generate smooth trajectories in the position and speed domains, thereby improving reconstruction accuracy. Rempe, et al. [89] present STRIVE, a method that leverages a conditional Varia-tional Autoencoder (VAE) with an expressive data-driven motion prior to creating challenging scenarios for stress-testing AV systems. Zhang, et al. [88] presented TrafficBots, which can effectively generate realistic multi-agent behaviors in densely populated urban environments, leveraging a shared, vectorized context and assigning each TrafficBot a personality learned using conditional variational autoencoder (CVAE). In multi-agent modeling, a frequently used formulation of Variational Autoencoders (VAEs) is vulnerable to social posterior collapse, a phenomenon in which the model tends to overlook the historical social context when predicting an agent's future trajectory. This can lead to significant prediction errors and suboptimal generalization performance [91]. Tang, et al. [91] developed a GNN-based framework, termed Social-CVAE, which incorporates an innovative Sparse-GAMP layer. This layer aids in detecting and analyzing social posterior collapse effectively. Jiao, et al. [90] propose a novel Trajectory Autoencoder (TAE) that incorporates drivers' behavior, such as aggressiveness and intention, into the latent space using a semi-supervised adversarial autoencoder and transportation domain knowledge.

3) GAN: GAN is a DL-based model comprising two neural networks: a generator and a discriminator. The generator network produces synthetic data, such as images, while the discriminator network assesses the authenticity of the generated data. During the training process, the generator learns to generate samples that are increasingly similar to the real data, while the discriminator becomes better at distinguishing between real and fake data. This adversarial relationship between the two networks leads to the generator producing increasingly realistic samples, while the discriminator becomes more accurate at identifying fake data. GAN allows one to learn more robust policies with fewer demonstrations, but adversarial training introduces another difficulty called mode collapse, where the generator produces only a limited number of samples that fail to capture the full diversity of the data. SimNet [67] uses conditional generative adversarial networks (cGANs) to initialize the state while ensuring consistency with the training distribution. Zhang, et al. [94] introduced D2Sim, a trajectory generation model that employs the Graph Attention Network with Transformer model and the Traffic Scenario Generation Information Maximizing GAN, which can effectively capture intricate and stochastic human driving behaviors derived from empirical data. Yin, et al. [95] propose RouteGAN, a generative model that generates diverse vehicle interactions by separately controlling their styles to produce trajectories with varying safety levels.

4) Diffusion: The diffusion model employs an Encoder-Decoder architecture, comprising a diffusion stage and an inverse diffusion stage. During the diffusion stage, the model progressively transforms the data from the original distribution to the desired distribution by iteratively adding noise to the original data, such as using Gaussian noise to convert the original data distribution to a normal distribution. In the inverse diffusion stage, a neural network is utilized to recover the data from the normal distribution to the original data distribution. Importantly, diffusion models allow a notion of control at generation time through so-called guidance, which has benefited several tasks. Zhong, et al. [96] proposed a controllable and realistic traffic simulation model that integrates the advantages of both heuristic-based and data-driven methodologies. By utilizing diffusion modeling and differentiable logic, they effectively steered generated trajectories to comply with rules established through signal temporal logic (STL). Pronovost, et al. [97] employed a unique combination of diffusion and object detection techniques to generate realistic and physically plausible configurations of discrete bounding boxes for agents, designed to simulate the output of a perception system in a self-driving car. Wang, et al. [98] proposed a world model called DriveDreamer, which can interpret complex environments, generate high-quality driving videos, and formulate realistic driving policies. They utilized Conditional Traffic Generation model (CTG) [96] model to construct a comprehensive representation of the complex environment.

*D. Other Deep Learning Models*

DL models employ neural networks to learn from vast amounts of data and make predictions. This type of artificial intelligence mimics the structure and function of the human brain. DL algorithms are designed to enhance their performance with increasing exposure to data. In traffic simulation, DL-based approaches are primarily utilized for two tasks: reactive simulation [29, 63, 84] and vision-based simulations from the FOV perspective [83, 85, 107-109].

1) Reactive simulation

Chang, et al. [84] treated VectorNet as a controller to simulate each agent's future behaviors and integrated a kinematic vehicle model into their framework to enhance closed-loop stability. However, this approach is vulnerable to distribution shifts. InterSim [63] resolved the issue of causal confusion by analyzing the interaction relationships between agents in the scene and producing realistic trajectories for every environment agent that aligns with these relationships. Leveraging graph neural networks (GNN) to combine the features of the agent and lane graphs, Suo, et al. [29] modeled agent goals as routes on the road network and trained a reactive route-conditional policy that explicitly incorporates these goals.

2) Vision-based simulations from the FOV perspective

Chen, et al. [85] presented a geometry-aware image composition process, GeoSim, which synthesizes novel urban driving scenarios by augmenting existing images with dynamic objects extracted from other scenes and rendered at novel poses. MetaSim [107] uses a graph neural network (GNN) to transform the attributes of each actor in the scene graph which has fewer constraints on possible generated scenes and can explicitly model a downstream task. Unfortunately, these approaches are still limited by their hand-crafted scene grammar which, for example, constrains vehicles to lane centerlines. Yang, et al. [83] developed UniSim, utilizing neural feature grids to reconstruct the static background and dynamic actors in the scene. It then composites them to generate LiDAR and camera data at novel viewpoints, with actors added or removed and at different locations. Currently, numerous researchers exploit NeRFs' realistic rendering capabilities for simulating autonomous driving scenarios. Utilizing training data obtained from real-world environments ensures a minimal sim-to-real gap. Wu, et al. [109] propose the integration of map priors into neural radiance fields (NeRF) for the synthesis of views outside the driving trajectory while maintaining semantic road consistency. Wu, et al. [108] introduced the initial open-source modular framework based on NeRF for simulating photorealistic autonomous driving, capable of managing both static and dynamic properties of instances.

## IV FUTURE WORK

Through the summary and research of the above content, we find several potential research directions that can shape the traffic simulation landscape. Here, we outline the identified futuristic research directions as follows:

- *Enhancing the robustness of models*: Current research in traffic simulation primarily focuses on IL, RL, and generative models, as well as DL. IL learns from demonstrations without interacting with the environment, which can lead to distribution shift and causal confusion, especially in long-term simulation. In contrast, RL allows the agent to learn by interacting with the environment and reward function, avoiding these issues. However, RL suffers from computational instability and difficulty in shaping the reward function. Generative models can generate long-tail scenarios by learning and capturing the underlying data distribution, but their computational complexity and resource-intensive nature are notable disadvantages. DL models, especially those with many parameters, are prone to overfitting, where they memorize the training data instead of generalizing to unseen data. Therefore, future work must address these problems to improve traffic simulation performance.

- *Improving the interpretability of models*: Most of the existing research uses neural networks to model agents, which are often considered as "black boxes" due to their intricate structures and nonlinear transformations. This lack of interpretability makes it difficult to understand the underlying decision-making processes and can hinder the identification of potential weaknesses or biases within the models. Game theory models have been widely used in the decision-making process of AVs, which can provide insights into the interactions between traffic participants and improve interpretability. However, their lack of universality limits their applicability to diverse traffic scenarios, due to the specific characteristics of game formulations. Liu, et al. [110] developed a three-level decision-making framework that uses a normal-form game to capture the interactions between the ego vehicle and surrounding vehicles. The framework includes a neural network designed analytically based on game principles, which enhances interpretability compared to existing data-driven approaches. However, this framework has not yet been extended to multi-vehicle scenarios. Future work can combine with other methods, such as game theory, or modularize the agents to enhance the interpretability of generating trajectories or behavior and interactions.

- *Large language models*: In the domain of artificial intelligence (AI), considerable interest has been garnered by large language models (LLMs). The application of reinforcement learning with human feedback (RLHF) has exemplified its scalability and promise as an approach to fine-tuning LLMs [74, 111]. The introduction of RLHF in the field of autonomous driving simulation presents a promising approach to enhance the realism and complexity of generated traffic scenarios. By incorporating human input and expertise into the learning process, RLHF enables the creation of more realistic and diverse situations that closely resemble real-world driving environments.

## V CONCLUSIONS

Given the safety concerns and high costs associated with real-world vehicle testing, autopilot simulation has become a widely utilized tool for rapid iteration and verification of driving algorithms [8]. Consequently, there is significant interest from researchers and industries in developing high-quality simulators. The current research focus is primarily on enhancing the realism and diversity of simulated scenarios by learning methods. We conducted a comprehensive investigation of data-driven traffic simulations for the first time. This survey paper not only contributes to a better understanding of autonomous driving simulation but also serves as a guideline for future research in the field. We have conducted a clear modularization of each part of the traffic simulation problem, from input modalities to modeling and finally output evaluation. In order to facilitate further research and development in this field, we have compiled a publicly available summary table of datasets. Subsequently, a classification of simulation models based on different methods was performed. Finally, this paper presents an overview of promising research directions in the field of traffic simulation, highlighting the necessity for models that are more robust and interpretable, as well as the incorporation of emerging technologies. The survey aims to offer valuable insights for both researchers and practitioners in this field and to steer the future development of traffic simulation.

## REFERENCES

[1] M. Zhu, X. Wang, and J. Hu, "Impact on car following behavior of a forward collision warning system with headway monitoring," *Transportation research part C: emerging technologies,* vol. 111, pp. 226-244, 2020.

[2] M. Zhu, "Behavior Modeling and Motion Planning for Autonomous Driving using Artificial Intelligence," 2022.

[3] M. Zhu, X. Wang, and Y. Wang, "Human-like autonomous car-following model with deep reinforcement learning," *Transportation research part C: emerging technologies,* vol. 97, pp. 348-368, 2018.

[4] J. Wang, L. Zhang, D. Zhang, and K. Li, "An adaptive longitudinal driving assistance system based on driver characteristics," *IEEE Transactions on Intelligent Transportation Systems,* vol. 14, no. 1, pp. 1-12, 2012.

[5] Z. Huang, X. Xu, H. He, J. Tan, and Z. Sun, "Parameterized batch reinforcement learning for longitudinal control of autonomous land vehicles," *IEEE Transactions on Systems, Man, and Cybernetics: Systems,* vol. 49, no. 4, pp. 730-741, 2017.

[6] P. A. Lopez *et al.*, "Microscopic traffic simulation using sumo," in *2018 21st international conference on intelligent transportation systems (ITSC),* 2018: IEEE, pp. 2575-2582.

[7] M. Fellendorf and P. Vortisch, "Microscopic traffic flow simulator VISSIM," *Fundamentals of traffic simulation,* pp. 63-93, 2010.

[8] A. Dosovitskiy, G. Ros, F. Codevilla, A. Lopez, and V. Koltun, "CARLA: An open urban driving simulator," in *Conference on robot learning,* 2017: PMLR, pp. 1-16.

[9] M. Treiber, A. Hennecke, and D. Helbing, "Congested traffic states in empirical observations and microscopic simulations," *Physical review E,* vol. 62, no. 2, p. 1805, 2000.

[10] V. Bharilya and N. Kumar, "Machine Learning for Autonomous Vehicle's Trajectory Prediction: A comprehensive survey, Challenges, and Future Research Directions," *arXiv preprint arXiv:2307.07527,* 2023.

[11] Y. Ma *et al.*, "Vision-centric bev perception: A survey," *arXiv preprint arXiv:2208.02797,* 2022.

[12] X. Li, J. Li, X. Hu, and J. Yang, "Line-cnn: End-to-end traffic line detection with line proposal unit," *IEEE Transactions on Intelligent Transportation Systems,* vol. 21, no. 1, pp. 248-258, 2019.

[13] B. Zhou and P. Krähenbühl, "Cross-view transformers for real-time map-view semantic segmentation," in *Proceedings of the IEEE/CVF conference on computer vision and pattern recognition*, 2022, pp. 13760-13769.

[14] Y. Liu, T. Yuan, Y. Wang, Y. Wang, and H. Zhao, "Vectormapnet: End-to-end vectorized hd map learning," in *International Conference on Machine Learning*, 2023: PMLR, pp. 22352-22369.

[15] J. Gao *et al.*, "Vectornet: Encoding hd maps and agent dynamics from vectorized representation," in *Proceedings of the IEEE/CVF Conference on Computer Vision and Pattern Recognition*, 2020, pp. 11525-11533.

[16] K. Guo, W. Jing, L. Gao, W. Liu, W. Li, and J. Pan, "Long-term Microscopic Traffic Simulation with History-Masked Multi-agent Imitation Learning," *arXiv preprint arXiv:2306.06401,* 2023.

[17] H. Zhou and J. Laval, "Longitudinal motion planning for autonomous vehicles and its impact on congestion: A survey," 2019.

[18] C. Laugier *et al.*, "Probabilistic analysis of dynamic scenes and collision risks assessment to improve driving safety," *IEEE Intelligent Transportation Systems Magazine,* vol. 3, no. 4, pp. 4-19, 2011.

[19] Q. Tran and J. Firl, "Online maneuver recognition and multimodal trajectory prediction for intersection assistance using non-parametric regression," in *2014 ieee intelligent vehicles symposium proceedings*, 2014: IEEE, pp. 918-923.

[20] G. Agamennoni, J. I. Nieto, and E. M. Nebot, "Estimation of multivehicle dynamics by considering contextual information," *IEEE Transactions on robotics,* vol. 28, no. 4, pp. 855-870, 2012.

[21] A. Eidehall and L. Petersson, "Statistical threat assessment for general road scenes using Monte Carlo sampling," *IEEE Transactions on intelligent transportation systems,* vol. 9, no. 1, pp. 137-147, 2008.

[22] J. Chen, C. Zhang, J. Luo, J. Xie, and Y. Wan, "Driving maneuvers prediction based autonomous driving control by deep Monte Carlo tree search," *IEEE transactions on vehicular technology,* vol. 69, no. 7, pp. 7146-7158, 2020.

[23] J. Wiest, M. Höffken, U. Kreßel, and K. Dietmayer, "Probabilistic trajectory prediction with Gaussian mixture models," in *2012 IEEE Intelligent vehicles symposium*, 2012: IEEE, pp. 141-146.

[24] G. S. Aoude, V. R. Desaraju, L. H. Stephens, and J. P. How, "Driver behavior classification at intersections and validation on large naturalistic data set," *IEEE Transactions on Intelligent Transportation Systems,* vol. 13, no. 2, pp. 724-736, 2012.

[25] G. Zheng, H. Liu, K. Xu, and Z. Li, "Objective-aware traffic simulation via inverse reinforcement learning," *arXiv preprint arXiv:2105.09560,* 2021.

[26] J. Mercat, T. Gilles, N. El Zoghby, G. Sandou, D. Beauvois, and G. P. Gil, "Multi-head attention for multi-modal joint vehicle motion forecasting," in *2020 IEEE International Conference on Robotics and Automation (ICRA)*, 2020: IEEE, pp. 9638-9644.

[27] N. Rhinehart, R. McAllister, K. Kitani, and S. Levine, "Precog: Prediction conditioned on goals in visual multi-agent settings," in *Proceedings of the IEEE/CVF International Conference on Computer Vision*, 2019, pp. 2821-2830.

[28] B. Varadarajan *et al.*, "Multipath++: Efficient information fusion and trajectory aggregation for behavior prediction," in *2022 International Conference on Robotics and Automation (ICRA)*, 2022: IEEE, pp. 7814-7821.

[29] S. Suo *et al.*, "MixSim: A Hierarchical Framework for Mixed Reality Traffic Simulation," in *Proceedings of the IEEE/CVF Conference on Computer Vision and Pattern Recognition*, 2023, pp. 9622-9631.

[30] P. Karkus, B. Ivanovic, S. Mannor, and M. Pavone, "Diffstack: A differentiable and modular control stack for autonomous vehicles," in *Conference on Robot Learning*, 2023: PMLR, pp. 2170-2180.

[31] S. Ettinger *et al.*, "Large scale interactive motion forecasting for autonomous driving: The waymo open motion dataset," in *Proceedings of the IEEE/CVF International Conference on Computer Vision*, 2021, pp. 9710-9719.

[32] B. Wilson *et al.*, "Argoverse 2: Next generation datasets for self-driving perception and forecasting," *arXiv preprint arXiv:2301.00493,* 2023.

[33] J. Bock, R. Krajewski, T. Moers, S. Runde, L. Vater, and L. Eckstein, "The ind dataset: A drone dataset of naturalistic road user trajectories at german intersections," in *2020 IEEE Intelligent Vehicles Symposium (IV)*, 2020: IEEE, pp. 1929-1934.

[34] R. Krajewski, T. Moers, J. Bock, L. Vater, and L. Eckstein, "The round dataset: A drone dataset of road user trajectories at roundabouts in germany," in *2020 IEEE 23rd International Conference on Intelligent Transportation Systems (ITSC)*, 2020: IEEE, pp. 1-6.

[35] W. Zhan *et al.*, "Interaction dataset: An international, adversarial and cooperative motion dataset in interactive driving scenarios with semantic maps," *arXiv preprint arXiv:1910.03088,* 2019.

[36] J. Houston *et al.*, "One thousand and one hours: Self-driving motion prediction dataset," in *Conference on Robot Learning*, 2021: PMLR, pp. 409-418.

[37] R. Krajewski, J. Bock, L. Kloeker, and L. Eckstein, "The highd dataset: A drone dataset of naturalistic vehicle trajectories on german highways for validation of highly automated driving systems," in *2018 21st international conference on intelligent transportation systems (ITSC)*, 2018: IEEE, pp. 2118-2125.

[38] V. G. Kovvali, V. Alexiadis, and L. Zhang PE, "Video-based vehicle trajectory data collection," 2007.

[39] P. Sun *et al.*, "Scalability in perception for autonomous driving: Waymo open dataset," in *Proceedings of the IEEE/CVF conference on computer vision and pattern recognition*, 2020, pp. 2446-2454.

[40] L. Chen *et al.*, "Persformer: 3d lane detection via perspective transformer and the openlane benchmark," in *European Conference on Computer Vision*, 2022: Springer, pp. 550-567.

[41] F. Yan *et al.*, "Once-3dlanes: Building monocular 3d lane detection," in *Proceedings of the IEEE/CVF Conference on Computer Vision and Pattern Recognition*, 2022, pp. 17143-17152.

[42] P. Xiao *et al.*, "Pandaset: Advanced sensor suite dataset for autonomous driving," in *2021 IEEE International Intelligent Transportation Systems Conference (ITSC)*, 2021: IEEE, pp. 3095-3101.

[43] J. Mao *et al.*, "One million scenes for autonomous driving: Once dataset," *arXiv preprint arXiv:2106.11037,* 2021.

[44] J. Geyer *et al.*, "A2d2: Audi autonomous driving dataset," *arXiv preprint arXiv:2004.06320,* 2020.

[45] N. Gählert, N. Jourdan, M. Cordts, U. Franke, and J. Denzler, "Cityscapes 3d: Dataset and benchmark for 9 dof vehicle detection," *arXiv preprint arXiv:2006.07864,* 2020.

[46] Y. Liao, J. Xie, and A. Geiger, "KITTI-360: A novel dataset and benchmarks for urban scene understanding in 2d and 3d," *IEEE Transactions on Pattern Analysis and Machine Intelligence,* vol. 45, no. 3, pp. 3292-3310, 2022.

[47] V. Ramanishka, Y.-T. Chen, T. Misu, and K. Saenko, "Toward driving scene understanding: A dataset for learning driver behavior and causal reasoning," in *Proceedings of the IEEE Conference on Computer Vision and Pattern Recognition*, 2018, pp. 7699-7707.

[48] M.-F. Chang *et al.*, "Argoverse: 3d tracking and forecasting with rich maps," in *Proceedings of the IEEE/CVF conference on computer vision and pattern recognition*, 2019, pp. 8748-8757.

[49] H. Caesar *et al.*, "nuscenes: A multimodal dataset for autonomous driving," in *Proceedings of the IEEE/CVF conference on computer vision and pattern recognition*, 2020, pp. 11621-11631.

[50] W. Maddern, G. Pascoe, C. Linegar, and P. Newman, "1 year, 1000 km: The oxford robotcar dataset," *The International Journal of Robotics Research,* vol. 36, no. 1, pp. 3-15, 2017.

[51] A. Patil, S. Malla, H. Gang, and Y.-T. Chen, "The h3d dataset for full-surround 3d multi-object detection and tracking in crowded urban scenes," in *2019 International Conference on Robotics and Automation (ICRA)*, 2019: IEEE, pp. 9552-9557.

[52] Q.-H. Pham *et al.*, "A 3D dataset: Towards autonomous driving in challenging environments," in *2020 IEEE International conference on Robotics and Automation (ICRA)*, 2020: IEEE, pp. 2267-2273.

[53] X. Huang, P. Wang, X. Cheng, D. Zhou, Q. Geng, and R. Yang, "The apolloscape open dataset for autonomous driving and its application," *IEEE transactions on pattern analysis and machine intelligence,* vol. 42, no. 10, pp. 2702-2719, 2019.

[54] F. Yu *et al.*, "Bdd100k: A diverse driving dataset for heterogeneous multitask learning," in *Proceedings of the IEEE/CVF conference on computer vision and pattern recognition*, 2020, pp. 2636-2645.

[55] A. Geiger, P. Lenz, C. Stiller, and R. Urtasun, "Vision meets robotics: The kitti dataset," *The International Journal of Robotics Research,* vol. 32, no. 11, pp. 1231-1237, 2013.

[56] S. Casas, C. Gulino, S. Suo, K. Luo, R. Liao, and R. Urtasun, "Implicit latent variable model for scene-consistent motion forecasting," in *Computer Vision–ECCV 2020: 16th European Conference, Glasgow, UK, August 23–28, 2020, Proceedings, Part XXIII 16*, 2020: Springer, pp. 624-641.

[57] S. Casas, C. Gulino, R. Liao, and R. Urtasun, "Spagnn: Spatially-aware graph neural networks for relational behavior forecasting from sensor data," in *2020 IEEE International Conference on Robotics and Automation (ICRA)*, 2020: IEEE, pp. 9491-9497.

[58] N. Kamra, H. Zhu, D. K. Trivedi, M. Zhang, and Y. Liu, "Multi-agent trajectory prediction with fuzzy query attention," *Advances in Neural Information Processing Systems*, vol. 33, pp. 22530-22541, 2020.

[59] S. Suo, S. Regalado, S. Casas, and R. Urtasun, "Trafficsim: Learning to simulate realistic multi-agent behaviors," in *Proceedings of the IEEE/CVF Conference on Computer Vision and Pattern Recognition*, 2021, pp. 10400-10409.

[60] D. Lee, Y. Gu, J. Hoang, and M. Marchetti-Bowick, "Joint interaction and trajectory prediction for autonomous driving using graph neural networks," *arXiv preprint arXiv:1912.07882*, 2019.

[61] Q. Sun, X. Huang, J. Gu, B. C. Williams, and H. Zhao, "M2i: From factored marginal trajectory prediction to interactive prediction," in *Proceedings of the IEEE/CVF Conference on Computer Vision and Pattern Recognition*, 2022, pp. 6543-6552.

[62] S. Kumar, Y. Gu, J. Hoang, G. C. Haynes, and M. Marchetti-Bowick, "Interaction-based trajectory prediction over a hybrid traffic graph," in *2021 IEEE/RSJ International Conference on Intelligent Robots and Systems (IROS)*, 2021: IEEE, pp. 5530-5535.

[63] Q. Sun, X. Huang, B. C. Williams, and H. Zhao, "InterSim: Interactive traffic simulation via explicit relation modeling," in *2022 IEEE/RSJ International Conference on Intelligent Robots and Systems (IROS)*, 2022: IEEE, pp. 11416-11423.

[64] M. Bansal, A. Krizhevsky, and A. Ogale, "Chauffeurnet: Learning to drive by imitating the best and synthesizing the worst," *arXiv preprint arXiv:1812.03079*, 2018.

[65] H. Caesar *et al.*, "nuplan: A closed-loop ml-based planning benchmark for autonomous vehicles," *arXiv preprint arXiv:2106.11810*, 2021.

[66] D. Xu, Y. Chen, B. Ivanovic, and M. Pavone, "BITS: Bi-level imitation for traffic simulation," in *2023 IEEE International Conference on Robotics and Automation (ICRA)*, 2023: IEEE, pp. 2929-2936.

[67] L. Bergamini *et al.*, "Simnet: Learning reactive self-driving simulations from real-world observations," in *2021 IEEE International Conference on Robotics and Automation (ICRA)*, 2021: IEEE, pp. 5119-5125.

[68] G. Zheng, H. Liu, K. Xu, and Z. Li, "Learning to simulate vehicle trajectories from demonstrations," in *2020 IEEE 36th International Conference on Data Engineering (ICDE)*, 2020: IEEE, pp. 1822-1825.

[69] X. Yan, Z. Zou, S. Feng, H. Zhu, H. Sun, and H. X. Liu, "Learning naturalistic driving environment with statistical realism," *Nature Communications*, vol. 14, no. 1, p. 2037, 2023.

[70] Z. Zhu *et al.*, "RITA: Boost Autonomous Driving Simulators with Realistic Interactive Traffic Flow," *arXiv preprint arXiv:2211.03408*, 2022.

[71] R. Bhattacharyya *et al.*, "Modeling human driving behavior through generative adversarial imitation learning," *IEEE Transactions on Intelligent Transportation Systems*, vol. 24, no. 3, pp. 2874-2887, 2022.

[72] F. Behbahani *et al.*, "Learning from demonstration in the wild," in *2019 International Conference on Robotics and Automation (ICRA)*, 2019: IEEE, pp. 775-781.

[73] R. P. Bhattacharyya, D. J. Phillips, B. Wulfe, J. Morton, A. Kuefler, and M. J. Kochenderfer, "Multi-agent imitation learning for driving simulation. In 2018 IEEE," in *RSJ International Conference on Intelligent Robots and Systems (IROS)*, pp. 1534-1539.

[74] Y. Cao, B. Ivanovic, C. Xiao, and M. Pavone, "Reinforcement Learning with Human Feedback for Realistic Traffic Simulation," *arXiv preprint arXiv:2309.00709*, 2023.

[75] M. Sackmann, H. Bey, U. Hofmann, and J. Thielecke, "Modeling Driver Behavior using Adversarial Inverse Reinforcement Learning," in *IEEE Intelligent Vehicles Symposium, Proceedings*, 2022, vol. 2022-June, pp. 1683-1690, doi: 10.1109/IV51971.2022.9827292.

[76] H. Niu *et al.*, "(Re) $^∧$ 2$ H2O: Autonomous Driving Scenario Generation via Reversely Regularized Hybrid Offline-and-Online Reinforcement Learning," *arXiv preprint arXiv:2302.13726*, 2023.

[77] B. Chen, X. Chen, Q. Wu, and L. Li, "Adversarial evaluation of autonomous vehicles in lane-change scenarios," *IEEE transactions on intelligent transportation systems*, vol. 23, no. 8, pp. 10333-10342, 2021.

[78] Q. Zhang *et al.*, "TrajGEN: Generating realistic and diverse trajectories with reactive and feasible agent behaviors for autonomous driving," *IEEE Transactions on Intelligent Transportation Systems*, vol. 23, no. 12, pp. 24474-24487, 2022.

[79] A. Mavrogiannis, R. Chandra, and D. Manocha, "B-gap: Behavior-rich simulation and navigation for autonomous driving," *IEEE Robotics and Automation Letters*, vol. 7, no. 2, pp. 4718-4725, 2022.

[80] P. Kothari, C. Perone, L. Bergamini, A. Alahi, and P. Ondruska, "Drivergym: Democratising reinforcement learning for autonomous driving," *arXiv preprint arXiv:2111.06889*, 2021.

[81] S. Feng, X. Yan, H. Sun, Y. Feng, and H. X. Liu, "Intelligent driving intelligence test for autonomous vehicles with naturalistic and adversarial environment," *Nature communications*, vol. 12, no. 1, p. 748, 2021.

[82] S. Shiroshita *et al.*, "Behaviorally diverse traffic simulation via reinforcement learning," in *2020 IEEE/RSJ International Conference on Intelligent Robots and Systems (IROS)*, 2020: IEEE, pp. 2103-2110.

[83] Z. Yang *et al.*, "UniSim: A Neural Closed-Loop Sensor Simulator," in *Proceedings of the IEEE/CVF Conference on Computer Vision and Pattern Recognition*, 2023, pp. 1389-1399.

[84] W.-J. Chang, Y. Hu, C. Li, W. Zhan, and M. Tomizaka, "Analyzing and Enhancing Closed-loop Stability in Reactive Simulation," in *2022 IEEE 25th International Conference on Intelligent Transportation Systems (ITSC)*, 2022: IEEE, pp. 3665-3672.

[85] Y. Chen *et al.*, "Geosim: Realistic video simulation via geometry-aware composition for self-driving," in *Proceedings of the IEEE/CVF conference on computer vision and pattern recognition*, 2021, pp. 7230-7240.

[86] L. Feng, Q. Li, Z. Peng, S. Tan, and B. Zhou, "Trafficgen: Learning to generate diverse and realistic traffic scenarios," in *2023 IEEE International Conference on Robotics and Automation (ICRA)*, 2023: IEEE, pp. 3567-3575.

[87] S. Tan, K. Wong, S. Wang, S. Manivasagam, M. Ren, and R. Urtasun, "Scenegen: Learning to generate realistic traffic scenes," in *Proceedings of the IEEE/CVF Conference on Computer Vision and Pattern Recognition*, 2021, pp. 892-901.

[88] Z. Zhang, A. Liniger, D. Dai, F. Yu, and L. Van Gool, "TrafficBots: Towards World Models for Autonomous Driving Simulation and Motion Prediction," *arXiv preprint arXiv:2303.04116*, 2023.

[89] D. Rempe, J. Philion, L. J. Guibas, S. Fidler, and O. Litany, "Generating useful accident-prone driving scenarios via a learned traffic prior," in *Proceedings of the IEEE/CVF Conference on Computer Vision and Pattern Recognition*, 2022, pp. 17305-17315.

[90] R. Jiao, X. Liu, B. Zheng, D. Liang, and Q. Zhu, "Tae: A semi-supervised controllable behavior-aware trajectory generator and predictor," in *2022 IEEE/RSJ International Conference on Intelligent Robots and Systems (IROS)*, 2022: IEEE, pp. 12534-12541.

[91] C. Tang, W. Zhan, and M. Tomizuka, "Exploring social posterior collapse in variational autoencoder for interaction modeling," *Advances in Neural Information Processing Systems*, vol. 34, pp. 8481-8494, 2021.

[92] R. Krajewski, T. Moers, A. Meister, and L. Eckstein, "BézierVAE: Improved trajectory modeling using variational autoencoders for the safety validation of highly automated vehicles," in *2019 IEEE Intelligent Transportation Systems Conference (ITSC)*, 2019: IEEE, pp. 3788-3795.

[93] R. Krajewski, T. Moers, D. Nerger, and L. Eckstein, "Data-driven maneuver modeling using generative adversarial networks and variational autoencoders for safety validation of highly automated vehicles," in *2018 21st International Conference on Intelligent Transportation Systems (ITSC)*, 2018: IEEE, pp. 2383-2390.

[94] K. Zhang, C. Chang, W. Zhong, S. Li, Z. Li, and L. Li, "A systematic solution of human driving behavior modeling and simulation for automated vehicle studies," *IEEE transactions on intelligent transportation systems*, vol. 23, no. 11, pp. 21944-21958, 2022.

[95] Z.-H. Yin, L. Sun, L. Sun, M. Tomizuka, and W. Zhan, "Diverse critical interaction generation for planning and planner evaluation," in *2021 IEEE/RSJ International Conference on Intelligent Robots and Systems (IROS)*, 2021: IEEE, pp. 7036-7043.

[96] Z. Zhong *et al.*, "Guided conditional diffusion for controllable traffic simulation," in *2023 IEEE International Conference on Robotics and Automation (ICRA)*, 2023: IEEE, pp. 3560-3566.

[97] E. Pronovost, K. Wang, and N. Roy, "Generating Driving Scenes with Diffusion," *arXiv preprint arXiv:2305.18452*, 2023.

[98] X. Wang, Z. Zhu, G. Huang, X. Chen, and J. Lu, "DriveDreamer: Towards Real-world-driven World Models for Autonomous Driving," *arXiv preprint arXiv:2309.09777*, 2023.

[99] S. Ross, G. Gordon, and D. Bagnell, "A reduction of imitation learning and structured prediction to no-regret online learning," in *Proceedings of the fourteenth international conference on artificial intelligence and statistics*, 2011: JMLR Workshop and Conference Proceedings, pp. 627-635.

[100] P. Isola, J.-Y. Zhu, T. Zhou, and A. A. Efros, "Image-to-image translation with conditional adversarial networks," in *Proceedings of the IEEE conference on computer vision and pattern recognition*, 2017, pp. 1125-1134.

[101] K. He, X. Zhang, S. Ren, and J. Sun, "Deep residual learning for image recognition," in *Proceedings of the IEEE conference on computer vision and pattern recognition*, 2016, pp. 770-778.

[102] M. Igl *et al.*, "Symphony: Learning realistic and diverse agents for autonomous driving simulation," in *2022 International Conference on Robotics and Automation (ICRA)*, 2022: IEEE, pp. 2445-2451.

[103] J. Devlin, M.-W. Chang, K. Lee, and K. Toutanova, "Bert: Pre-training of deep bidirectional transformers for language understanding," *arXiv preprint arXiv:1810.04805*, 2018.

[104] J. Fu, K. Luo, and S. Levine, "Learning robust rewards with adversarial inverse reinforcement learning," *arXiv preprint arXiv:1710.11248*, 2017.

[105] R. Chandra, U. Bhattacharya, T. Mittal, A. Bera, and D. Manocha, "Cmetric: A driving behavior measure using centrality functions," in *2020 IEEE/RSJ International Conference on Intelligent Robots and Systems (IROS)*, 2020: IEEE, pp. 2035-2042.

[106] M. Sackmann, H. Bey, U. Hofmann, and J. Thielecke, "Modeling driver behavior using adversarial inverse reinforcement learning," in *2022 IEEE Intelligent Vehicles Symposium (IV)*, 2022: IEEE, pp. 1683-1690.

[107] A. Kar *et al.*, "Meta-sim: Learning to generate synthetic datasets," in *Proceedings of the IEEE/CVF International Conference on Computer Vision*, 2019, pp. 4551-4560.

[108] Z. Wu *et al.*, "Mars: An instance-aware, modular and realistic simulator for autonomous driving," *arXiv preprint arXiv:2307.15058*, 2023.

[109] C. Wu, J. Sun, Z. Shen, and L. Zhang, "MapNeRF: Incorporating Map Priors into Neural Radiance Fields for Driving View Simulation," *arXiv preprint arXiv:2307.14981*, 2023.

[110] M. Liu, Y. Wan, F. L. Lewis, S. Nageshrao, and D. Filev, "A three-level game-theoretic decision-making framework for autonomous vehicles," *IEEE Transactions on Intelligent Transportation Systems*, vol. 23, no. 11, pp. 20298-20308, 2022.

[111] D. M. Ziegler *et al.*, "Fine-tuning language models from human preferences," *arXiv preprint arXiv:1909.08593*, 2019.

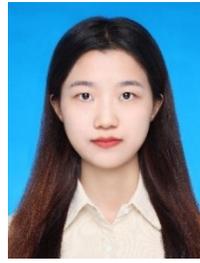

**Di Chen** received the B.Sc. and M.S. degree in Transportation Engineering from School of Transportation engineering, Tongji University, Shanghai, China, 2020. Currently, she is a research assistant in the Thrust of Intelligent Transportation under the Systems Hub at the Hong Kong University of Science and Technology (Guangzhou). Her research interests include agent-based simulation, machine learning methods for autonomous vehicles.

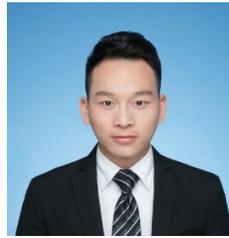

**Meixin Zhu** is a tenure-track Assistant Professor in the Thrust of Intelligent Transportation (INTR) under the Systems Hub at the Hong Kong University of Science and Technology (Guangzhou) and an affiliated Assistant Professor in the Civil and Environmental Engineering Department at the Hong Kong University of Science and Technology. He is also with Guangdong Provincial Key Lab of Integrated Communication, Sensing and Computation for Ubiquitous Internet of Things. He obtained a Ph.D. degree in intelligent transportation at the University of Washington (UW) in 2022. He received his BS and MS degrees in traffic engineering in 2015 and 2018, respectively, from Tongji University. His research interests include Autonomous Driving Decision Making and Planning, Driving Behavior Modeling, Traffic-Flow Modeling and Simulation, Traffic Signal Control, and (Multi-Agent) Reinforcement Learning. He is a recipient of the TRB Best Dissertation Award (AED50) in 2023.

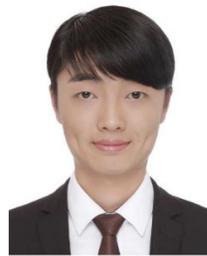

**Hao (Frank) Yang** received the joint B.S. degree in telecommunication engineering from the Beijing University of Posts and Telecommunications and the University of London in 2017, and the Ph.D. degree with the Smart Transportation Research and Application Laboratory (STAR Lab) in the Department of Civil and Environmental Engineering at University of Washington. He will be joining the Department of Civil & System Engineering (CaSE) at Johns Hopkins University as a tenure-track assistant professor in Fall 2024. He is an Associate Editor for IEEE ITSC 2021 and a reviewer of CVPR, IEEE TRANSACTIONS ON INTELLIGENT TRANSPORTATION SYSTEMS, and IEEE TRANSACTIONS ON NEURAL NETWORKS AND LEARNING SYSTEMS. His research interests include computer vision, edge computing, and transportation data analysis.

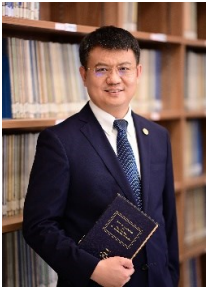

**Xuesong Wang** received the Ph.D. degree from the University of Central Florida, Orlando, FL, USA, in 2006. In Fall 2008, he joined Tongji University, Shanghai, China, where he is currently a Professor at the School of Transportation Engineering. His research interests include safety assessment of autonomous vehicles, naturalistic driving study, driving behavior analysis, traffic safety analysis, safety evaluation of roadway design, and transportation safety planning. On these topics, he has authored or coauthored over 350 papers in journals and conferences. He is a member of the TRB Standing Committees on Transportation in the Developing Countries (AME40), Safety Performance Analysis (ACS20), and Road User Measurement and Evaluation (ACH50). He is a Handing Editor of Transportation Research Record, an Associate Editor of Accident Analysis and Prevention, and an Editorial Board Member of the Journal of Transportation Safety and Security.

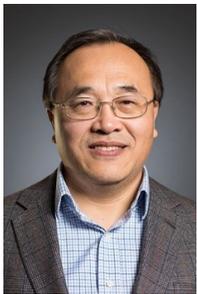

**Yinhai Wang** received the master's degree in computer science from the University of Washington (UW) and the Ph.D. degree in transportation engineering from The University of Tokyo in 1998. He is currently a professor in transportation engineering and the Founding Director of the Smart Transportation Applications and Research Laboratory (STAR Lab), UW. He also serves as the Director of the Pacific Northwest Transportation Consortium (PacTrans), U.S. Department of Transportation, University Transportation Center for Federal Region 10.